\documentclass{article}

\usepackage[preprint]{neurips_2025}

\usepackage{hyperref}       
\usepackage[utf8]{inputenc} 
\usepackage[T1]{fontenc}    
\usepackage{url}            
\usepackage{booktabs}       
\usepackage{amsfonts}       
\usepackage{nicefrac}       
\usepackage{microtype}      
\usepackage{xcolor}         
\usepackage{graphicx}
\usepackage{enumitem}
\usepackage{multirow} 
\usepackage{adjustbox}
\usepackage{subcaption}
\usepackage{wrapfig}
\usepackage{url}
\usepackage{amsmath}
\usepackage[capitalize]{cleveref}

\definecolor{denim}{rgb}{0.08, 0.38, 0.74}
\hypersetup{
  colorlinks   = true, 
  urlcolor     = magenta, 
  linkcolor    = denim, 
  citecolor   =  denim,
}

\title{UPCORE: Utility-Preserving Coreset Selection for Balanced Unlearning}

\author{%
  Vaidehi Patil \\
  \And
Elias Stengel-Eskin\\
\\
  UNC Chapel Hill \\
  \And
  Mohit Bansal 
}

\newcommand{\myparagraph}[1]{\textbf{#1}\hspace{0.4em}}
\newcommand{\ourmethod}{\textsc{UPCORE}}

\begin{document}

\maketitle

\begin{abstract}
User specifications or legal frameworks often require information to be removed from pretrained models, including large language models (LLMs). This requires deleting or ``forgetting'' a set of data points from an already-trained model, which typically degrades its performance on other data points. Thus, a balance must be struck between removing information and keeping the model's other abilities intact, with a failure to balance this trade-off leading to poor deletion or an unusable model. 
To this end, we propose \ourmethod{} (Utility-Preserving Coreset Selection), a method-agnostic data selection framework for mitigating collateral damage during unlearning. 
Finding that the model damage is correlated with the variance of the model's representations on the forget set, we selectively prune the forget set to remove outliers, thereby minimizing model degradation after unlearning. Across three standard unlearning methods, \ourmethod{} consistently achieves a superior balance between the competing objectives of deletion efficacy and model preservation. To better evaluate this trade-off, we introduce a new metric, measuring the area-under-the-curve (AUC) across standard metrics.
Our results show that \ourmethod{} improves both standard metrics and AUC, benefiting from positive transfer between the coreset and pruned points while reducing negative transfer from the forget set to points outside of it.\footnote{Correspondence to: Vaidehi Patil <vaidehi@cs.unc.edu>. Code and data: \url{https://github.com/Vaidehi99/UPCORE}\\}
\end{abstract}

\section{Introduction}
\label{sec:intro}

The widespread deployment of ML models, especially large language models (LLMs), has raised concerns around privacy, regulation, and ethical use. Trained on massive, uncurated web data, these models often memorize sensitive, copyrighted, or undesirable content \citep{shokri2017membership, carlini2019secret}. With regulations like the GDPR and CCPA granting individuals the ``right to be forgotten,'' efficient methods for removing specific data or topics from pre-trained models are increasingly necessary. Machine unlearning offers a promising solution by enabling targeted data removal without full retraining. Beyond regulatory compliance, it also helps reduce harmful outputs, protect intellectual property, and align LLMs with societal values \citep{jang2023knowledge}. These challenges have driven renewed interest in improving model editing and unlearning techniques \citep{liu2024rethinking, hase2024fundamental}.

\begin{figure}[ht]
    \centering
    \includegraphics[width=0.92\linewidth]{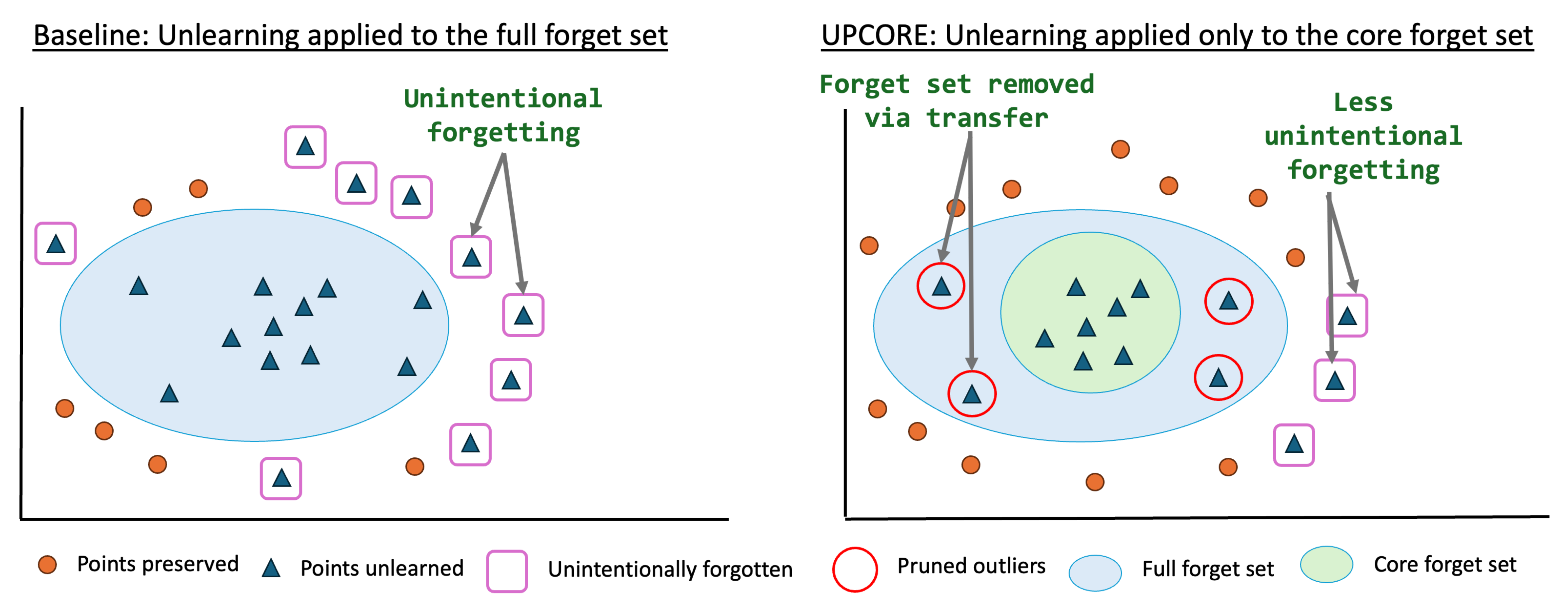}
    \vspace{-0.5em}
    \caption{\textbf{Left:} Standard unlearning methods are applied equally to all points in the forget set. 
    Here, outlier points in the model's hidden space (visualized in 2D) contribute to the unintentional forgetting of points outside of the forget set (i.e. collateral damage).
    \textbf{Right:} By finding a lower-variance coreset within the forget set, \ourmethod{} reduces damage while maintaining forget performance via positive transfer from the coreset to the pruned points.
    }
    \label{fig:intro}
\end{figure}

Given the growing use of LLMs, prior work has proposed methods for removing knowledge or skills \citep{cao2015towards, bourtoule2021machine, nguyen2022survey} and steering behavior in targeted ways \citep{sinitsin2020editable, meng2022locating}. 
However, such editing often induces unintended side effects, reducing utility on unrelated tasks. 
Effective unlearning therefore requires balancing deletion of undesired information with preservation of overall model utility. 
To evaluate this, current methods assess both deletion success (via ``forget set'' performance) and collateral effects on unrelated behaviors (via ``retain set'' accuracy). 
This is particularly important in realistic, topic-level unlearning scenarios—such as removing information about an individual or sensitive domain \citep{li2024the}—where deletion may cause over-generalization and degrade performance on semantically or structurally similar inputs.

A key gap in existing research lies in understanding the specific data characteristics that drive over-generalization and collateral effects during unlearning. 
While prior work \citep{sheshadri2024latent, chowdhury2024towards} has measured damage resulting from unlearning,
it does not investigate how attributes of the data -- such as its variance -- contribute to collateral damage or whether these attributes can be controlled to optimize the trade-off between deletion efficacy and utility retention. 
Focusing on a topic-based setting where the forget set comprises semantically coherent groups of information, we seek to address these questions:

\begin{enumerate}[itemsep=0.0em, leftmargin=*]
    \item {\it What measurable attributes of the forget set drive collateral effects during the unlearning process?}
    \item {\it Can these attributes be systematically controlled to optimize the trade-off between deletion effectiveness and model utility?}
\end{enumerate}

We investigate which properties of the forget data correlate with collateral damage during unlearning. 
Our analysis reveals a strong positive correlation between the variance of the model's hidden states corresponding to datapoints in the forget set (hidden state variance, or HSV), and the extent of collateral damage to the model after unlearning.
In other words, unlearning a set of widely-distributed datapoints (as shown in \cref{fig:intro} (left)) leads to more damage than unlearning a more densely-distributed set. 
Building on this insight, we hypothesize that selectively curating a coreset with lower variance from the larger forget set can help optimize this trade-off, as shown in \cref{fig:intro} (right).

To this end, we introduce \ourmethod{}, which constructs a core forget set by systematically identifying and pruning data points in the forget set that contribute most to the variance and thereby to collateral damage. \ourmethod{} organizes points into an Isolation Forest \citep{IsolationForest}, which identifies anomalous points in a set. 
By pruning these points, we reduce the variance within the forget set, which we find leads to less damage. 
Crucially, in addition to reducing collateral damage, \ourmethod{} in fact leverages it by identifying two separate kinds of collateral effects: (1) \textbf{Negative collateral damage}: Unintended degradation of unrelated model capabilities and (2) \textbf{Positive collateral transfer}: The intended impact on pruned data points removed to form the core forget set. 
This is illustrated in \cref{fig:intro}, where pruned outlier points are still unlearned -- despite not being a part of the coreset used for unlearning -- due to positive transfer, and is further highlighted by our results in \cref{tab:results} and \cref{tab:rouge}, which show that \ourmethod{} results in better unlearning than a randomly-selected subset while also having better knowledge retention on non-forget data. We show positive collateral transfer enabled by \ourmethod{} in \cref{tab:positive}, with deletion transferring to points outside the coreset.
Moreover, our focus on data makes \ourmethod{} method-agnostic: it can be applied to any data-driven unlearning framework.

We evaluate \ourmethod{} in prompt completion and question-answering settings and across three standard unlearning methods: Gradient Ascent \citep{jang2023knowledge}, Refusal \citep{ouyang2022training, maini2024tofu} and Negative Preference Optimization (NPO) \citep{zhang2024negative}, applying each unlearning algorithm directly to the optimized core forget set obtained using \ourmethod{}, rather than the entire forget set.
We measure three critical dimensions: (1)
\emph{Unlearning effectiveness}, measured by the successful removal of targeted knowledge in the (a) forget set, (b) paraphrased versions of removed information as well as (c) prompts attempting to jailbreak the model.; (2) \textit{Unintended damage}, where we quantify collateral effects on unrelated model capabilities; and (3) \textit{Intended transfer}, where we analyze the impact on the pruned data points that were removed from the core forget set as shown \cref{sec:positive_transfer}. 
While we follow past unlearning work in the metrics we use to measure the trade-off between the competing objectives, we also note that the current suite of metrics measures performance at a fixed point during unlearning.
This can make comparisons across methods hard, as the trade-off between deletion efficacy and model utility varies across unlearning gradient update steps.
To address this, in addition to showing improvements on standard metrics, we introduce a novel set of metrics that report the area-under-the-curve (AUC) for the standard unlearning metric suite, reporting not just the performance at one fixed timestep, but measuring \emph{how a method trades off deletion with model utility across checkpoints (see \cref{fig:auc_comparison}).}

Empirically, we find that across all three unlearning methods, \ourmethod{} consistently has the highest AUC compared to baselines of unlearning on the complete forget set and choosing a random subset of forget points.
In other words, \ourmethod{} forms a Pareto frontier, maximizing unlearning effectiveness while also minimizing model damage.
Moreover, \ourmethod{} positively leverages generalization by transferring unlearning from the core set to the high-variance outlier points that were removed from the core forget set.
Notably, it consistently beats baselines across \emph{all} unlearning methods; this holds true across multiple metrics (e.g. ROUGE on a ``retain'' set, on neighborhood data closely related to (but not in) the forget set, etc.). \ourmethod{}'s superior trade-off effectively generalizes to variations of the forgotten information, performing well on paraphrased versions of forgotten prompts as well as prompts intended to jailbreak the model.
We also see these gains reflected in static evaluations of one checkpoint (as opposed to AUC, which evaluates across checkpoints); here, \ourmethod{} obtains lower (better) ROUGE on the forget set than the random baseline while simultaneously incurring less model damage than the random and complete baselines, with the best (highest) ROUGE across all data not in the forget set. 

\vspace{-0.5em}
\section{Background and Related Work}
\vspace{-0.5em}

\myparagraph{Unlearning Methods for LLMs.} Machine unlearning methods fall into two categories: \emph{exact unlearning}, which ensures the model is indistinguishable from one retrained without the forget data, and \emph{approximate unlearning}, which modifies model parameters efficiently without full retraining. Due to the high cost of retraining LLMs, most approaches—including ours—use the latter. One class of methods trains models via RLHF to produce uninformative responses (e.g., ``I don't know'') on forget prompts \citep{ouyang2022training, wen2024know}. \citet{yao2023large} apply gradient ascent to suppress harmful outputs, substituting them with whitespace, though this causes utility loss on benign prompts. To mitigate such degradation, \citet{chen2023unlearn} introduce an unlearning layer effective across tasks, while \citet{eldan2023s} propose a specialized architecture for removing copyrighted content. For evaluation, \citet{maini2024tofu} present a benchmark we adopt. Despite these advances, managing the trade-off between unlearning and utility remains challenging. We address this by introducing a data-driven coreset selection framework to minimize collateral damage.

\myparagraph{Model Editing for Unlearning.} Model editing provides an alternative approach to unlearning by directly modifying model weights to forget target facts \citep{de-cao-etal-2021-editing, dai2022knowledge, mitchell2022fast, meng2022locating}. 
Following model editing work \citep{patil2024unlearning}, our framework employs LoRA updates with standard unlearning objectives (See \cref{app:editing} for more details).

\myparagraph{Coreset Selection.} Coreset selection identifies representative subsets that preserve key dataset properties, improving computational efficiency. Given the NP-hard complexity of the exhaustive search, methods have focused on optimizing coverage, diversity, or importance \citep{sener2018active, NEURIPS2023_3abe23bf}. By recognizing unequal contributions of data points, coreset selection has proven effective in supervised learning \citep{wei2015submodularity, killamsetty2021glister, killamsetty2021grad}, enabling efficient performance. 
Our work forms new connections between these methods and the problem of unlearning in LLMs, where preserving utility and minimizing collateral damage are critical.

\section{Methods}

We introduce \ourmethod{} (Utility-Preserving Coreset Design for Unlearning),
an approach motivated by the observation that certain data points in the forget set disproportionately contribute to collateral damage during unlearning, primarily by increasing data variance. 
To address this, \ourmethod{} reformulates pruning for core forget set selection as an outlier detection task, where outlier data points i.e. points with the greatest influence on utility degradation are identified and pruned. 
By minimizing variance within the forget set, \ourmethod{} reduces unintended negative effects, ensuring more effective and targeted unlearning.

\subsection{Problem Definition}
\begin{figure*}
    \centering
    \includegraphics[width=\textwidth]{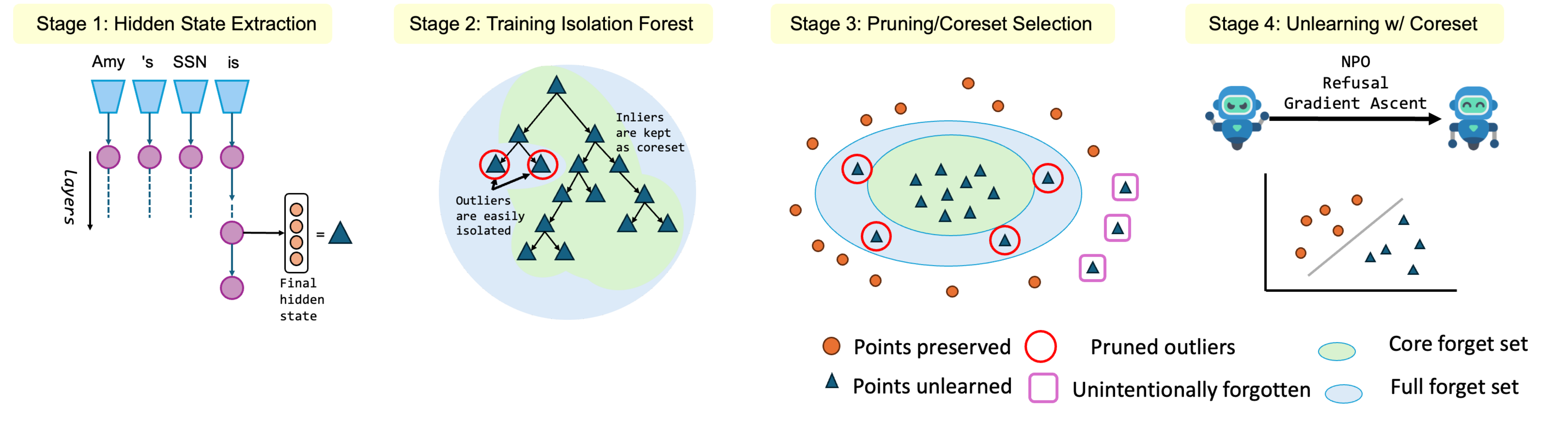}
    \vspace{-1.5em}
    \caption{\ourmethod{} has four stages. First, we extract hidden states from the LLM to be modified; second, we identify outliers using Isolation Forests; third, we prune outliers to select a core forget set, and fourth, we perform unlearning on the coreset.}
    \label{fig:method}
\end{figure*}

Let \( D \) be the dataset used to train a model \( M \), with \( D_F \subset D \) representing the forget set to be unlearned. 
Directly unlearning \( D_F \), i.e., applying an unlearning algorithm \( \mathcal{U} \) to the model \( M \) using only the data points in \( D_F \), produces an updated model \( M' = \mathcal{U}(M, D_F) \). However, this often leads to significant performance degradation on the retained dataset \( D \setminus D_F \) due to over-generalization, where unlearning updates undesirably propagate beyond the forget set, impacting unrelated data points in $D \setminus D_F$.

To address this issue, we aim to construct a pruned forget set \( D_C \subset D_F \), henceforth referred to as the core forget set, by removing points in \( D_F \) that disproportionately drive over-generalization. 
The goal is to balance two competing objectives: (i) minimizing negative collateral damage, i.e., performance degradation on \( D \setminus D_F \), and (ii) maintaining deletion accuracy, i.e., ensuring that the unlearning \( D_C \) effectively deletes the undesirable knowledge associated with the original forget set \( D_F \). 
More formally, given a damage metric \( \text{Damage}_{(\mathcal{U}, D_C)}(M, M', D \setminus D_F) \) that quantifies the impact of unlearning $D_C$ on \( D \setminus D_F \), and a deletion accuracy metric \( \text{DelAcc}_{(\mathcal{U}, D_C)}(M', D_F) \) that evaluates the effectiveness of \( \mathcal{U} \) in forgetting \( D_F \) after unlearning on \( D_C \), the problem can be formulated as an optimization task: $D_C = \arg\min_{D_C \subseteq D_F} \bigg( 
 \text{Damage}_{(\mathcal{U}, D_C)}\big(M, M', (D \setminus D_F)\big) - \lambda \cdot \text{DelAcc}_{(\mathcal{U}, D_C)}\big(M', D_F\big) 
\bigg)$ where \( \lambda > 0 \) is a hyperparameter that controls the trade-off between the competing objectives.

\subsection{Variance as a Measure of Collateral Damage} 
\label{sec:analysis}

Building on prior work analyzing the cross-task generalization of forgetting methods \citep{zhang2024unforgettable}, we investigate the relationship between attributes of the forget set \( D_F \) and their impact on collateral damage during unlearning i.e. \(\text{Damage}_{(\mathcal{U}, D_C)}\big(M, M', (D \setminus D_F)\big)\). Specifically, we identify the variance \(Var( D_F )\) as a critical predictor of overgeneralization. Since directly selecting optimal subsets based on unlearning performance is computationally infeasible -- due to the high cost of repeatedly retraining or unlearning -- our goal is to develop scalable heuristics like variance that approximate which subsets minimize collateral damage, enabling efficient and practical coreset selection for unlearning.
To systematically evaluate the relationship between variance and unlearning, we analyze question-answer (QA) pairs generated from Wikipedia documents across diverse topics.
Each topic-specific dataset acts as the forget set \( D_F \), with unlearning applied separately, one topic at a time.
For each forget set, we compute variance using the hidden states of the last token and the penultimate layer of the model. 
We then compute retain set performance as the model utility metric proposed by \citet{maini2024tofu}, which measures performance on preserved data points after unlearning. 
The results, visualized in \cref{fig:hsv} in \cref{append:hsv_analysis}, demonstrate a strong negative correlation between HSV and model utility, indicating that variance is a potential driver of over-generalization. 
These findings underscore the importance of identifying and excluding points that lead to higher variance i.e. outliers to mitigate utility loss. 
In \cref{append:hsv_analysis} we show similar analyses for other attributes such as model confidence and gradient similarity but find no strong correlation between utility degradation and these attributes.

\subsection{\ourmethod{}: Core Forget Set Selection}

To achieve variance minimization in the forget set \( D_F \), \ourmethod{} frames the problem as an outlier detection task.  
\ourmethod{} provides two key benefits: (1) It mitigates negative collateral damage by pruning outliers to form a more compact core forget set, and (2) It strategically exploits collateral over-generalization to extend unlearning beyond the core forget set, effectively removing the pruned points as well.
As shown in \cref{fig:intro}, what might traditionally be viewed as detrimental collateral damage -- when it affects points outside the forget set ($D_F$) -- can be turned to our advantage when it impacts untrained data points within the forget set that were pruned (\( D_F \setminus D_C\)).

To detect these outliers, we use the Isolation Forest algorithm \citep{IsolationForest}, an unsupervised learning technique that efficiently identifies anomalous data points. 
Isolation Forest works by recursively partitioning the dataset using random feature selections and random split values. 
Points that are isolated with fewer partitions, i.e. are isolated more easily, are considered outliers, as they differ significantly from the majority of the data. 
This makes the Isolation Forest algorithm particularly effective for high-dimensional data where traditional distance-based outlier detection methods may fail. 
These outliers, characterized by their isolation in the feature space, are likely to contribute to high variance and over-generalization during the unlearning process.
\ourmethod{} proceeds as follows (illustrated in \cref{fig:method}):

  {\bf Stage 1: Hidden Feature Extraction:} We extract hidden state representations \( \mathcal{H} \) from the model's penultimate layer, as it typically encodes high-level semantic abstractions while retaining generalization capacity, unlike the final layer which is often biased toward task-specific outputs \citep{skean2025layer} 
  (See \cref{fig:method} left), corresponding to the final token of each question in \( D_F \). These representations, which reflect the model's internal representation of the data, serve as input features for outlier detection. This step is guided by our analysis in \cref{sec:analysis}, which highlights the strong link between hidden state variance and collateral damage. 

  {\bf Stage 2: Training the Isolation Forest and Computing Anomaly Scores:}  
  We train an Isolation Forest model \( \mathcal{I} \) on the forget set \( D_F \) to model its distribution, recursively partitioning the data to detect outliers (see \cref{fig:method} middle). Points isolated more quickly and requiring fewer splits are flagged as outliers, indicating disproportionate contributions to variance in the hidden state space. For each \( d \in D_F \), \( \mathcal{I} \) assigns an anomaly score \( \text{score}(d) \) based on the average path length \( h(d) \) required to isolate the point across an ensemble of binary trees. Shorter path lengths correspond to higher anomaly scores, indicating points that contribute to variance and thus collateral effects. 
  Additional details are provided in \cref{sec:appendix}.

\textbf{Stage 3: Prune Outliers and Setting Stopping Criterion:} To construct the pruned coreset \( D_C \), we apply a threshold \( \tau \) on the anomaly scores from the Isolation Forest model: $D_C = \{ d \in D_F \mid \text{score}(d) \leq \tau \}$. 
Data points with scores above \( \tau \) are excluded as outliers, as they disproportionately contribute to variance. 
We hypothesize that removing these outliers will reduce utility degradation while preserving core information for forgetting. The threshold \( \tau \) is determined via a stopping criterion, which can be chosen as follows: (1) Coreset Size Control: Specify a desired coreset size \( |D_C| \) to ensure an appropriate number of inliers (2) Proportional Pruning: Select the top \( k\% \) of points with the lowest anomaly scores to maintain a consistent pruning ratio. By selecting \( \tau \) based on user requirements, \ourmethod{} provides fine-grained control over the trade-off. In practice, we prune 10\% of the data points in our main experiments and additionally conduct scaling experiments that vary the pruning percentage (see \cref{sec:coreset_size}) to analyze its impact on the trade-off dynamics.

{\bf Stage 4: Unlearning on the Coreset:} After selecting the pruned coreset \( D_C \), \ourmethod{} applies the unlearning algorithm \( \mathcal{U} \) to the model \( M \), resulting in \( M'_{\ourmethod{}} = \mathcal{U}(M, D_C) \) (See \cref{fig:method} end). This process removes the influence of \( D_F \) while minimizing utility degradation on \( D \setminus D_F \). By focusing on \( D_C \), \ourmethod{} ensures targeted unlearning and  positively leverages collateral effects, as unlearning \( D_C \) also deletes much of \( D_F \)'s influence, even those parts not explicitly included in 
\( D_C \). We later show this positive transfer empirically in \cref{tab:positive}.

\section{Experimental Setup}
\label{sec:setup}
\myparagraph{Unlearning Methods and Baselines.}
We test \ourmethod{} with three standard unlearning methods: gradient ascent, refusal, and negative preference optimization, applied to a Llama-3.1-8B \citep{dubey2024llama} base model.
In all cases, models are trained using both the forget set (complete or sampled) and a retain set, which contains examples of data that should \emph{not} be forgotten, providing a contrastive signal. 
We consider the following unlearning methods:
\begin{itemize}[noitemsep,topsep=0pt,leftmargin=*]
\item \textbf{Gradient Ascent} \citep{jang2023knowledge}: Gradient Ascent \emph{maximizes} the training loss on the forget set \( D_F \). For each \( x \in D_F \), the objective is to maximize the loss.

\item \textbf{Refusal} \citep{ouyang2022training}: Refusal trains the model to respond to sensitive prompts with neutral, non-informative answers, such as ``\emph{I don't know.}''
     
\item \textbf{Negative Preference Optimization (NPO)} \citep{zhang2024negative}: NPO is a stable form of DPO \citep{rafailov2024direct} designed for unlearning. It reduces the gap between the likelihood of the target data and the likelihood from the original model while ensuring the unlearned model remains closely aligned with the original (See \cref{app:npo} for more details). 
\end{itemize}

We evaluate \ourmethod{}, which is a dataset selection method, against three other selection methods as baselines: (1) unlearning applied to the entire forget set (i.e. \emph{no selection}), and (2) unlearning performed on a randomly subsampled subset of the forget set, matched in size to the coreset curated by \ourmethod{} (i.e. \emph{random selection}) (3) $D^2$-pruning \citep{maharana2024mathbbd}, a standard coreset selection method that selects the coreset based on example diversity and difficulty.

We evaluate on factual questions across two settings: {\it prompt completion} and {\it question answering}, described below, and we include further details on these settings in \cref{append:data_detail}.

\subsection{Metrics and Answer Extraction}
Following prior work \citep{maini2024tofu}, we evaluate models using a suite of metrics. We compute ROUGE \citep{lin2004rouge} between reference and model answers to assess both utility and deletion effectiveness, as ROUGE captures content overlap in factual QA, unlike classification-based metrics with fixed labels. To measure unintended model damage, we report ROUGE on the retain set, neighborhood data, and Real World / Real Authors datasets \citep{maini2024tofu}, where higher is better. For deletion, we compute ROUGE on the forget set (lower is better) and on pruned forget examples to assess positive collateral transfer. We also report model utility, defined as the harmonic mean of the normalized conditional probabilities \( P(a \mid q)^{1/|a|} \), following \citet{cho-etal-2014-properties}, and the truth ratio, which compares the likelihood of correct vs.\ incorrect answers \citep{maini2024tofu}. We also evaluate all metrics on paraphrased and jailbreak variants of the forget set, where paraphrased variants reword target examples \citep{krishna2023paraphrasing} and jailbreak variants probe adversarial prompts designed to bypass unlearning \citep{zou2023universal, jin2024guard}.

\myparagraph{AUC Metric.}

While ROUGE and model utility are standard unlearning metrics, they provide only a single-point snapshot of model performance. This is limiting, as unlearning inherently involves a tradeoff between forgetting and model damage that evolves over training steps.\footnote{We use steps throughout; in main results, 1 epoch = 50 steps.} Early checkpoints may retain higher utility but underperform on deletion, while later ones improve forgetting at the cost of increased damage (\cref{fig:auc_comparison}). Since the number of unlearning steps varies across works, direct comparisons become difficult. To address this, we propose evaluating unlearning as a dynamic tradeoff over time. Instead of reporting ROUGE Forget and ROUGE Retain at one step, we compute the \emph{area under the curve} (AUC) between these metrics over unlearning steps. This tradeoff is visualized in \cref{fig:auc_comparison}, which plots inverse ROUGE on forget data (X-axis) versus ROUGE on neighboring data (Y-axis).\footnote{This curve differs from \cref{tab:results}: we compute AUC before averaging across topics here, and after in \cref{tab:results}.} We construct Pareto curves comparing deletion metrics (e.g., ROUGE Forget) with utility metrics (e.g., ROUGE Retain, ROUGE Neighborhood) across steps, and use the AUC as a unified, global measure of performance. We also study the effect of evaluation granularity on AUC (\cref{tab:granularity}), and show that AUC correlates with overall unlearning effectiveness (\cref{tab:rouge}) and negatively with forget data variance (\cref{tab:correlation}).
In \cref{append:hsv_analysis} we analyze the effect of varying the number of steps on this AUC metric and find our results are stable across different granularities. 
Therefore we adopt a standard of 50 steps, corresponding to one epoch.

\section{Experimental Results and Discussion}
\label{sec:results}

\subsection{\ourmethod{} Balances Deletion and Model Utility}

\myparagraph{Design.} To assess whether \ourmethod{} improves the Pareto frontier between deletion effectiveness and utility retention, we compare AUC values of models trained with \ourmethod{} and baseline methods (\cref{sec:setup}). AUC is computed using: (1) \textit{Deletion Effectiveness}, measured as $(1 - \text{ROUGE})$ on the forget set (X-axis), and (2) \textit{Utility Retention}, measured via ROUGE on non-forget datasets, including neighborhood data and the aggregated model utility metric from \citet{maini2024tofu} (Y-axis). We perform evaluations are performed on both Counterfact and TriviaQA topics.

\begin{figure}[t]
    \centering
    \begin{minipage}[t]{0.45\textwidth}
        \centering
        \includegraphics[width=\linewidth]{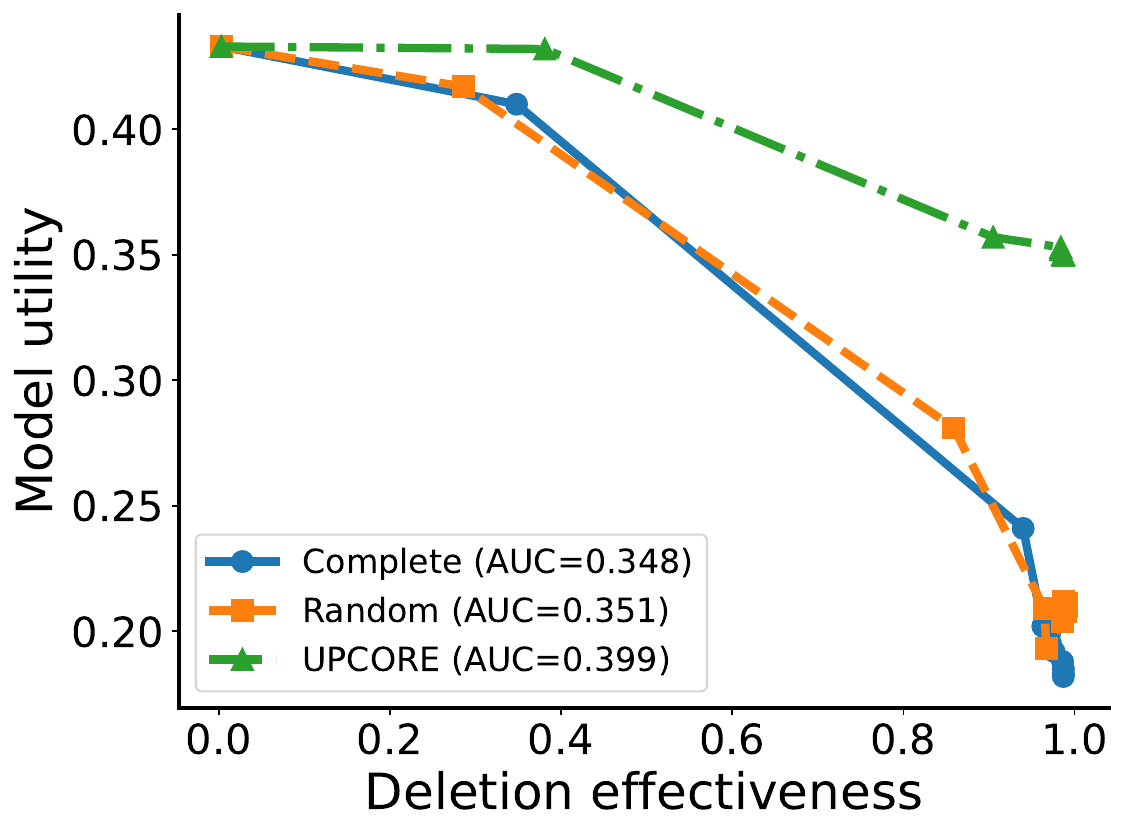}
        \vspace{-1.5em}
        \caption{Trading-off between deletion effectiveness and model utility forms a Pareto frontier across steps, shown here averaged across Counterfact topics with Gradient Ascent. 
        }
        \label{fig:auc_comparison}
    \end{minipage}%
    \hfill
    \begin{minipage}[t]{0.52\textwidth}
        \centering
        \includegraphics[width=\linewidth]{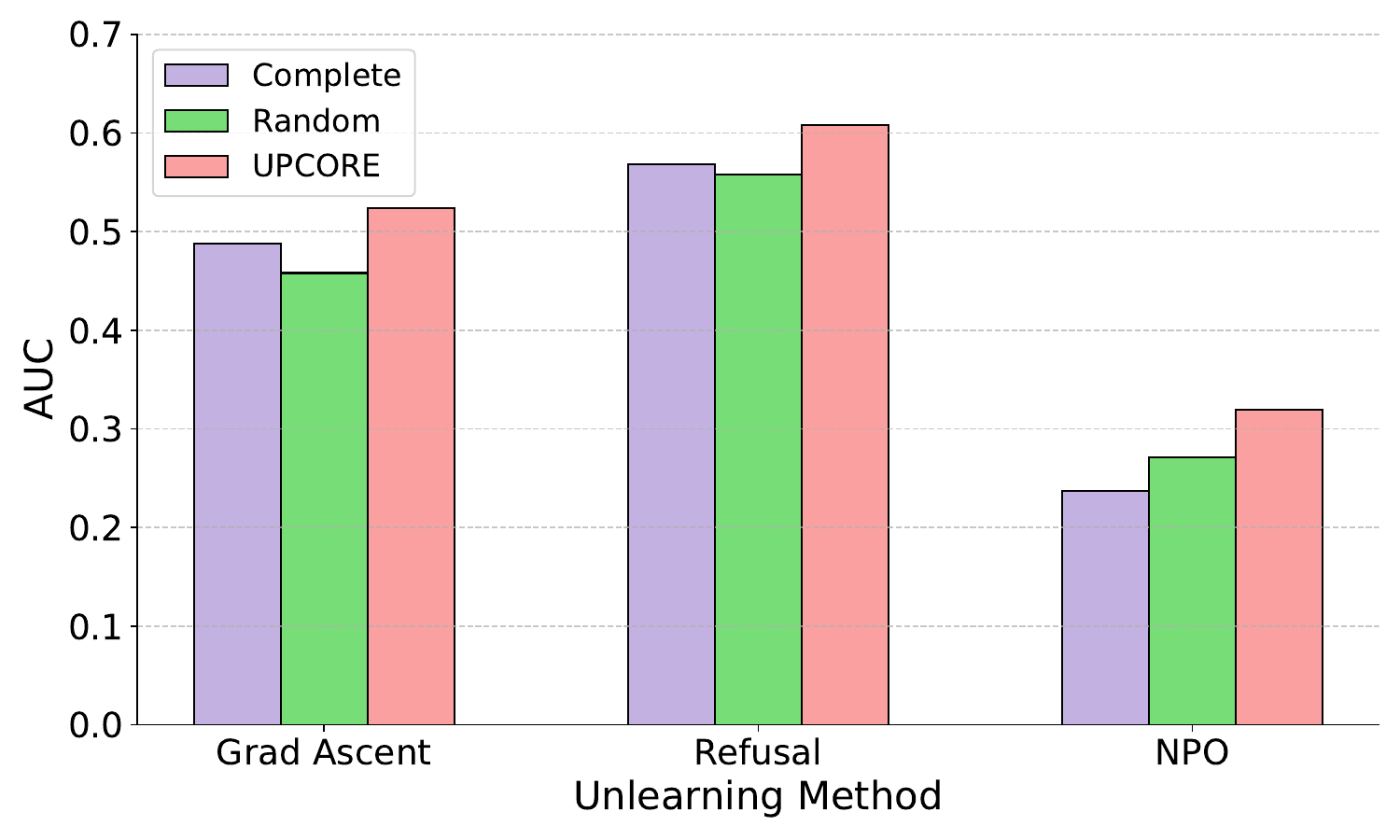}
        \vspace{-1.5em}
        \caption{AUC between forget set ROUGE and neighborhood data ROUGE averaged across topics in Counterfact. \ourmethod{} reduces damage to neighborhood data.
        }
        \label{fig:neighbor}
    \end{minipage}
\end{figure}

\myparagraph{Results.}
\cref{fig:auc_comparison} illustrates the AUC metric, which quantifies the trade-off between forget performance and model utility by measuring the area under the Pareto frontier. A higher AUC reflects better balance -- more effective forgetting with less utility loss. \ourmethod{} slows utility degradation by unlearning on a variance-based coreset. \cref{tab:results} shows that across the Counterfact dataset and three unlearning methods, \ourmethod{} consistently outperforms baselines -- complete forget set and random subsample -- by 3 to 7 AUC points, confirming its method-agnostic effectiveness. \cref{tab:results_trivia} extends this to TriviaQA (using gradient ascent) (see \cref{tab:results_trivia_2} for results using other unlearning methods on TriviaQA), where \ourmethod{} maintains its advantage with up to 3 AUC points improvement. Finally, \cref{tab:rouge} reports ROUGE and utility scores at epoch 10, showing that \ourmethod{} achieves the highest utility and non-forget ROUGE, while preserving competitive forget performance. We report p-values in Appendix \cref{tab:pvalues}, finding the gains are statistically significant.

\begin{table*}[t]
    \centering
    \caption{AUC across the two competing objectives: (1) \textit{Deletion Effectiveness}, defined as $(1 - \text{ROUGE})$ on the forget set (X-axis), and (2) \textit{Model Utility}, averaged across Counterfact topics and evaluated via ROUGE scores on multiple utility datasets, including neighborhood data and an aggregate model utility across datasets (Y-axis). We compare three unlearning methods: Gradient Ascent, Refusal, and NPO. Error bars indicate standard deviation across 3 seeds.}
    \vspace{0.5em}
  \begin{adjustbox}{width=\linewidth}
    \begin{tabular}{ccccccc}
    \toprule
        \textbf{Method} & \textbf{Selection} & \textbf{Retain} & \textbf{Neigh} & \textbf{Real World} & \textbf{Real Authors} & \textbf{Model Utility}\\ \midrule
        
        \multirow{3}{*}{Grad. Ascent} 
        & Complete & 0.488 {\scriptsize$\pm$ 0.015} & 0.568 {\scriptsize$\pm$ 0.018} & 0.720 {\scriptsize$\pm$ 0.016} & 0.891 {\scriptsize$\pm$ 0.020} & 0.343 {\scriptsize$\pm$ 0.012} \\ 
        & Random & 0.495 {\scriptsize$\pm$ 0.017} & 0.558 {\scriptsize$\pm$ 0.016} & 0.731 {\scriptsize$\pm$ 0.015} & 0.907 {\scriptsize$\pm$ 0.019} & 0.353 {\scriptsize$\pm$ 0.014} \\ 
        & $D^2$-pruning & 0.493 {\scriptsize$\pm$ 0.016} & 0.552 {\scriptsize$\pm$ 0.017} & 0.723 {\scriptsize$\pm$ 0.016} & 0.920 {\scriptsize$\pm$ 0.018} & 0.349 {\scriptsize$\pm$ 0.013} \\ 
        & \ourmethod{} & \textbf{0.523} {\scriptsize$\pm$ 0.008} & \textbf{0.608} {\scriptsize$\pm$ 0.010} & \textbf{0.769} {\scriptsize$\pm$ 0.009} & \textbf{0.933} {\scriptsize$\pm$ 0.011} & \textbf{0.387} {\scriptsize$\pm$ 0.007} \\ \midrule

        \multirow{3}{*}{Refusal} 
        & Complete & 0.493 {\scriptsize$\pm$ 0.016} & 0.488 {\scriptsize$\pm$ 0.017} & 0.714 {\scriptsize$\pm$ 0.015} & 0.890 {\scriptsize$\pm$ 0.018} & 0.366 {\scriptsize$\pm$ 0.014} \\ 
        & Random & 0.456 {\scriptsize$\pm$ 0.015} & 0.458 {\scriptsize$\pm$ 0.016} & 0.644 {\scriptsize$\pm$ 0.014} & 0.819 {\scriptsize$\pm$ 0.017} & 0.332 {\scriptsize$\pm$ 0.013} \\ 
        & $D^2$-pruning & 0.473 {\scriptsize$\pm$ 0.013} & 0.478 {\scriptsize$\pm$ 0.014} & 0.632 {\scriptsize$\pm$ 0.011} & 0.805 {\scriptsize$\pm$ 0.015} & 0.341 {\scriptsize$\pm$ 0.010} \\

        & \ourmethod{} & \textbf{0.500} {\scriptsize$\pm$ 0.007} & \textbf{0.524} {\scriptsize$\pm$ 0.009} & \textbf{0.744} {\scriptsize$\pm$ 0.008} & \textbf{0.920} {\scriptsize$\pm$ 0.010} & \textbf{0.381} {\scriptsize$\pm$ 0.006} \\ \midrule

        \multirow{3}{*}{NPO} 
        & Complete & 0.281 {\scriptsize$\pm$ 0.014} & 0.237 {\scriptsize$\pm$ 0.015} & 0.192 {\scriptsize$\pm$ 0.013} & 0.342 {\scriptsize$\pm$ 0.017} & 0.199 {\scriptsize$\pm$ 0.012} \\ 
        & Random & 0.253 {\scriptsize$\pm$ 0.015} & 0.271 {\scriptsize$\pm$ 0.014} & 0.195 {\scriptsize$\pm$ 0.013} & 0.308 {\scriptsize$\pm$ 0.016} & 0.186 {\scriptsize$\pm$ 0.011} \\ 
        & $D^2$-pruning & 0.265 {\scriptsize$\pm$ 0.013} & 0.254 {\scriptsize$\pm$ 0.014} & 0.193 {\scriptsize$\pm$ 0.012} & 0.320 {\scriptsize$\pm$ 0.016} & 0.190 {\scriptsize$\pm$ 0.011} \\
        & \ourmethod{} & \textbf{0.329} {\scriptsize$\pm$ 0.006} & \textbf{0.319} {\scriptsize$\pm$ 0.008} & \textbf{0.246} {\scriptsize$\pm$ 0.007} & \textbf{0.414} {\scriptsize$\pm$ 0.009} & \textbf{0.248} {\scriptsize$\pm$ 0.005} \\ 
    \bottomrule
    \end{tabular}
    \end{adjustbox}
    \label{tab:results}
    \vspace{-0.5em}
\end{table*}

\begin{table*}[t]
    \centering
    \caption{Evaluation metrics from \cref{tab:results} shown for Gradient Ascent on the {\bf TriviaQA} topics. Error bars indicate standard deviation across 3 seeds.}
    \vspace{0.5em}
    \begin{adjustbox}{width=\linewidth}
    \begin{tabular}{ccccccc}
    \toprule
        \textbf{Method} & \textbf{Selection} & \textbf{Retain} & \textbf{Neigh} & \textbf{Real World} & \textbf{Real Authors} & \textbf{Model Utility}\\ \midrule
        \multirow{3}{*}{Grad. Ascent} 
        & Complete & 0.153 {\scriptsize$\pm$ 0.004} & 0.285 {\scriptsize$\pm$ 0.005} & 0.226 {\scriptsize$\pm$ 0.004} & 0.155 {\scriptsize$\pm$ 0.003} & 0.135 {\scriptsize$\pm$ 0.004} \\ 
        & Random &  0.159 {\scriptsize$\pm$ 0.005} & 0.304 {\scriptsize$\pm$ 0.006} & 0.222 {\scriptsize$\pm$ 0.005} & 0.157 {\scriptsize$\pm$ 0.004} & 0.136 {\scriptsize$\pm$ 0.004} \\  
        & $D^2$-pruning & 0.162 {\scriptsize$\pm$ 0.003} & 0.310 {\scriptsize$\pm$ 0.004} & 0.224 {\scriptsize$\pm$ 0.003} & 0.157 {\scriptsize$\pm$ 0.003} & 0.141 {\scriptsize$\pm$ 0.003} \\
        & \ourmethod{} & \textbf{0.165} {\scriptsize$\pm$ 0.002} & \textbf{0.318} {\scriptsize$\pm$ 0.003} & \textbf{0.227} {\scriptsize$\pm$ 0.002} & \textbf{0.158} {\scriptsize$\pm$ 0.002} & \textbf{0.147} {\scriptsize$\pm$ 0.002} \\ 
    \bottomrule
    \end{tabular}
    \end{adjustbox}
        \label{tab:results_trivia}

\end{table*}

\begin{table*}[t]
    \centering
    \caption{ROUGE scores and model utility across topics from the Counterfact dataset for a fixed epoch of Gradient Ascent. 
    \ourmethod{} consistently has higher performance on data outside the forget set, with the least degradation among methods and closest performance to the base model, while still having a high forget rate.}
    \vspace{0.5em}
    \begin{tabular}{c|c|ccccc}
    \toprule
        
        \textbf{Method} & \textbf{Forget} & \textbf{Retain} & \textbf{Neigh.} & \textbf{Real Authors} & \textbf{Real World} & \textbf{Model Utility} \\ \midrule
        {\textcolor{gray}{\emph{Base model}}} & {\textcolor{gray}{\emph{0.997}}} & {\textcolor{gray}{\emph{0.546}}} & {\textcolor{gray}{\emph{0.820}}}  & {\textcolor{gray}{\emph{1.000}}} & {\textcolor{gray}{\emph{0.872}}} & {\textcolor{gray}{\emph{0.433}}} \\
        \hline
        Complete & 0.018 & 0.381 & 0.144  & 0.669 & 0.446 & 0.182 \\ 
        Random &  {\bf 0.011}  & 0.411 & 0.104 & \textbf{0.724} & 0.499 & 0.211 \\ 
        \ourmethod{} & 0.017 & \textbf{0.430} & {\bf 0.190}  & 0.706 & \textbf{0.528} & \textbf{0.350} \\ 
        \bottomrule
    \end{tabular}
    \label{tab:rouge}

\end{table*}

\subsection{Positive and Negative Transfer}
\label{sec:positive_transfer}

\myparagraph{Design.} Here, we measure both positive and negative transfer.
To assess whether unlearning on the core forget set induces deletion in the pruned data points (positive collateral transfer), we measure the ROUGE score of the unlearned model on these points. 
A significant drop in ROUGE would indicate that the forgetting process extends beyond the explicitly unlearned subset.
We measure negative transfer on the neighborhood data, examining the AUC between ROUGE on the neighborhood datapoints and forget set ROUGE. 

\begin{wraptable}{r}{0.45\textwidth}
    \centering
    \caption{ROUGE score on pruned datapoints. 
    Both for \ourmethod{} and random sampling, unlearning on a subset of datapoints translates to other datapoints not in the subset.}
    \vspace{0.5em}
    \begin{tabular}{ccc}
    \toprule
        \textbf{Method} & \textbf{Random} & \textbf{\ourmethod{}} \\ \midrule
        {\textcolor{gray}{\emph{Base model}}} & {\textcolor{gray}{\emph{1.000}}} & {\textcolor{gray}{\emph{1.000}}} \\ \hline
        Gradient Ascent & 0.022 & 0.053 \\ 
        Refusal         & 0.169 & 0.127 \\ 
        NPO             & 0.206 & 0.231 \\ 
    \bottomrule
    \end{tabular}
    \label{tab:positive}
    \vspace{-1em}
\end{wraptable}

\myparagraph{Results.} As shown in \cref{tab:positive}, ROUGE on pruned points drops from 1.00 to 0.053 (Gradient Ascent) and 0.127 (Refusal), indicating that unlearning transfers to pruned points despite not being directly targeted—likely due to topic-level over-generalization. This transfer is not unique to \ourmethod{}; a similar drop occurs with a random subsample of the same size, suggesting that pruned points share a semantic neighborhood with the forget set and are thus indirectly affected. However, in terms of negative transfer, \ourmethod{} achieves substantially higher AUC on neighborhood data (\cref{fig:neighbor}), indicating reduced unintended damage compared to random sampling. These findings are further supported by utility AUC gains in \cref{tab:results}, highlighting that while unlearning generalizes well across topics in the positive direction, variance-based pruning better limits collateral damage in the negative direction. We hypothesize that the gains on Counterfact are higher than those on TriviaQA (see \cref{tab:results} and \cref{tab:results_trivia}) due to the latter’s higher semantic density, which facilitates stronger positive transfer across related examples.

\subsection{Robustness to Jailbreaks}

\myparagraph{Design.}
To evaluate robustness against blackbox attacks, we test whether unlearning on the core forget set generalizes to adversarial/jailbreak prompts designed to elicit the same information (see \cref{append:data_examples} for examples and generation details). 
We report the AUC of (1-ROUGE) and ROUGE on non-forget data (e.g., retain set, neighborhood) in \cref{tab:jailbreak}. 
Higher AUC indicates greater robustness against extraction attacks \citep{zou2023universal,jin2024guard}. 
We repeat this analysis with paraphrased variants in Appendix \cref{tab:jailbreak_2}, with similar findings in terms of robustness to paraphrased queries.

\begin{table*}[t]
    \centering

    \caption{Evaluation metrics from \cref{tab:results} averaged across topics in Counterfact shown for Gradient Ascent, assessed for robustness to {\bf jailbreak} variants of the forget data with the same utility data.}
    \vspace{0.5em}
    \begin{tabular}{ccccccc}
    \toprule
    
        \textbf{Method} & \textbf{Selection} & \textbf{Retain} & \textbf{Neigh} & \textbf{Real World} & \textbf{Real Authors} & \textbf{Model Utility}\\ \midrule
        \multirow{3}{*}{Jailbreak} & Complete & 0.417 & 0.474 & 0.599 & 0.743 & 0.291\\ 
        & Random & 0.430 & 0.470 & 0.629 & 0.787 & 0.305\\ 
        & \ourmethod{} & {\bf 0.455} & {\bf 0.512} & {\bf 0.665} & {\bf 0.819} & {\bf 0.335} \\  
       \bottomrule
    \end{tabular}
    
    \label{tab:jailbreak}
\end{table*}

\myparagraph{Results.} As shown in \cref{tab:jailbreak} and \cref{tab:jailbreak_2}, \ourmethod{} achieves higher AUC across settings and utility datasets, outperforming baselines even under rephrases and jailbreak attacks. This indicates a superior trade-off and suggests that positive transfer from the core forget set generalizes to input variations eliciting the same target information.

\subsection{Correlation of Hidden-State Variance and Utility.} 
\label{append:hsv_analysis}
\begin{wrapfigure}{r}{0.35\textwidth}
        \centering
        \vspace{-4em}
        \includegraphics[width=\linewidth]{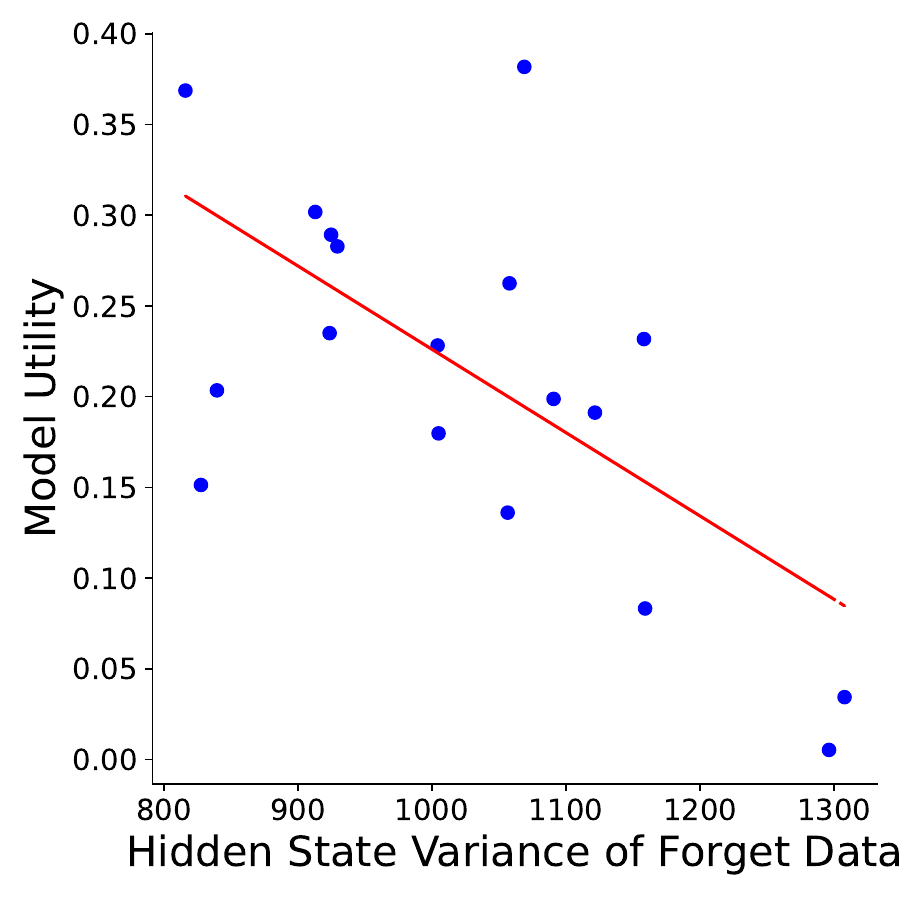}
        \caption{Model utility and hidden state variance (forget data) show a strong negative correlation (-0.714) across data from multiple topics.}
        \label{fig:hsv_main}
        \vspace{-4em}
\end{wrapfigure} 
We show that hidden state variance is strongly negatively correlated with model utility after unlearning (Pearson correlation = -0.714), supporting its role as a reliable proxy for unlearning-induced degradation. Additionally, \cref{append:auc_hsv} demonstrates a consistent inverse relationship between variance and AUC across topics, reinforcing that minimizing variance leads to improved deletion-utility trade-offs.
As illustrated in \cref{fig:hsv_main} (top left), we plot the hidden state variance of different data clusters against the utility metric from \citet{maini2024tofu} after a fixed number of unlearning steps, and observe a clear negative correlation.

\subsection{Additional Results Summary}

We briefly describe a number of additional results and analyses included in the appendix. 

two are strongly negatively correlated with a Pearson correlation of -0.714. 

\myparagraph{Alternate Outlier Detection Methods.} In \cref{append:outlier}, we empirically compare multiple outlier detection methods and find that Isolation Forest achieves the highest AUC, indicating its superior ability to identify informative outliers for pruning. 

\myparagraph{Effect of Forget Set Size.} \cref{append:forget_size} shows that \ourmethod{} retains its effectiveness even when the forget set is reduced to 50\% of its original size, maintaining improved utility relative to baselines; however, we hypothesize a lower bound below which pruning may become ineffective due to limited signal. 

\myparagraph{Granularity of AUC Steps.} \cref{append:granularity} examines the impact of varying the number of unlearning steps used to compute AUC and finds the results to be stable across different granularities. 
This motivates our choice of epochs for the main results in \cref{sec:results}.

\myparagraph{Coreset Size.} In \cref{sec:coreset_size} we examine the effect of scaling the size of the coreset, increasing the number of outliers pruned. 
We find that different pruning percentages generally lead to similar AUCs, as higher pruning leads to less model damage but also less forgetting.

\vspace{-0.5em}
\section{Conclusion}
\vspace{-0.5em}
We propose \ourmethod{}, a utility-preserving coreset selection framework for unlearning in LLMs that minimizes collateral damage while ensuring effective deletion. Empirically, we find that hidden state variance in the forget data strongly influences utility degradation. By pruning high-variance outliers to form a core forget set, \ourmethod{} improves the trade-off between deletion and retention. We quantify this trade-off using area-under-the-curve across unlearning steps. Results show that \ourmethod{} substantially reduces unintended performance loss and can be combined with any data-driven unlearning method, offering a principled and generalizable approach to utility-aware unlearning.

\section*{Acknowledgments}

We would like to thank Peter Hase for his feedback on an initial draft of the paper. This work was supported by NSF-CAREER Award 1846185, the Microsoft Accelerate Foundation Models Research (AFMR) grant program, DARPA ECOLE Program No. HR00112390060, and NSF-AI Engage Institute DRL-2112635. Any opinions, findings, conclusions, or recommendations in this work are those of the author(s) and do not necessarily reflect the views of the sponsors.

\bibliography{custom}
\bibliographystyle{plainnat}

\section{Additional Background}
\label{sec:appendix}

\subsection{Machine Unlearning Background.} 
The concept of machine unlearning \citep{cao2015towards} is typically divided into two categories: \emph{exact unlearning} and \emph{approximate unlearning}. 
Exact unlearning aims to completely remove information related to specific data, ensuring that the resulting model behaves identically to a model retrained from scratch without the forget data \citep{ginart2019making}. 
However, the computational infeasibility of retraining LLMs from scratch renders exact unlearning impractical for real-world applications. 
Approximate unlearning methods, on the other hand, focus on ensuring that the model parameters closely approximate those of a retrained model while maintaining computational efficiency \citep{guo2020certified, chien2022efficient, pan2023unlearning, yoon2025safree}.

\subsection{Coreset Selection.} 
Unlike prior work, which focuses on coreset selection for improving training efficiency or robustness, our approach leverages a novel perspective by applying coreset principles to the problem of machine unlearning. Specifically, while conventional methods \citep{maharana2024mathbbd} aim to preserve model accuracy during training by selecting representative data, our framework, \ourmethod{}, is designed to mitigate negative collateral damage during unlearning by identifying and pruning data points that disproportionately influence performance degradation. Furthermore, unlike general coreset selection approaches that primarily target classification or regression tasks \citep{Lee_2024_CVPR, wei2015submodularity}, our method is tailored for unlearning settings where the goal is retaining model utility while ensuring the effective removal of unwanted information. Thus, our work extends the applicability of coreset selection beyond traditional use cases, offering a principled approach to balancing unlearning effectiveness with model performance.

\subsection{Anomaly Score in Isolation Forest:} 
Isolation Forests produce anomaly scores for each point. 
More formally, the anomaly score for a data point \( d \) is defined as:

\[
\text{score}(d) = 2^{-\frac{h(d)}{c(n)}}
\]

where \( h(d) \) is the average path length for \( d \) across the ensemble of trees, \( n \) is the size of the dataset \( D_F \), and \( c(n) \) is the average path length for a dataset of size \( n \) in a random binary search tree. The term \( c(n) \) is given by:

\[
c(n) = 2H(n-1) - \frac{2(n-1)}{n}
\]

where \( H(i) \) denotes the \( i \)-th harmonic number, defined as \( H(i) = \sum_{j=1}^{i} \frac{1}{j} \).

\section{Method Details and Analysis}

\subsection{Scaling the Coreset Size} \label{sec:coreset_size}
\myparagraph{Design.} 
Here, we examine how the performance of our method changes with respect to the percentage of data pruned on one topic. 
Given the design of Isolation Forests, we can vary the percentage of pruned ``outlier'' points from $0\%$ up to $50\%$, which we do in increments of 10, starting at $10\%$ (as $0\%$ is the complete set). 
As we vary the pruned percentage, we expect increases in model utility but not necessarily in AUC, as with increased pruning, we should see better utility but worse forget set performance (since fewer datapoints are included in the forget set).

\myparagraph{Results.}

\cref{fig:scale} shows AUC scores across different coreset pruning percentages, averaged over topics from the Counterfact dataset.
\ourmethod{} achieves the largest performance gain between 0\% and 10\% pruning, followed by a dip at 20\%. Beyond 30\%, performance stabilizes across coreset sizes.
This trend reflects a core trade-off in coreset design: pruning more aggressively reduces the number of examples explicitly unlearned, which can weaken deletion effectiveness, but it also limits model damage, improving utility.
Interestingly, this plateau beyond 30\% suggests that positive transfer from the remaining examples can only compensate for deletion loss up to a point. Once that ceiling is reached, the competing forces—improved utility versus diminished forgetting—begin to balance out, resulting in the observed stability.

\subsection{\ourmethod{} Lowers Forget Set Variance}
\label{append:variance}

\paragraph{Design.} To verify that \ourmethod{} indeed leads to a lower variance compared to the random baseline, 
we report the hidden state variance of the forget set used in each baseline and in \ourmethod{}.

\begin{figure}[t]
    \centering
    \begin{minipage}[t]{0.48\textwidth}
        \centering
        \includegraphics[width=\linewidth]{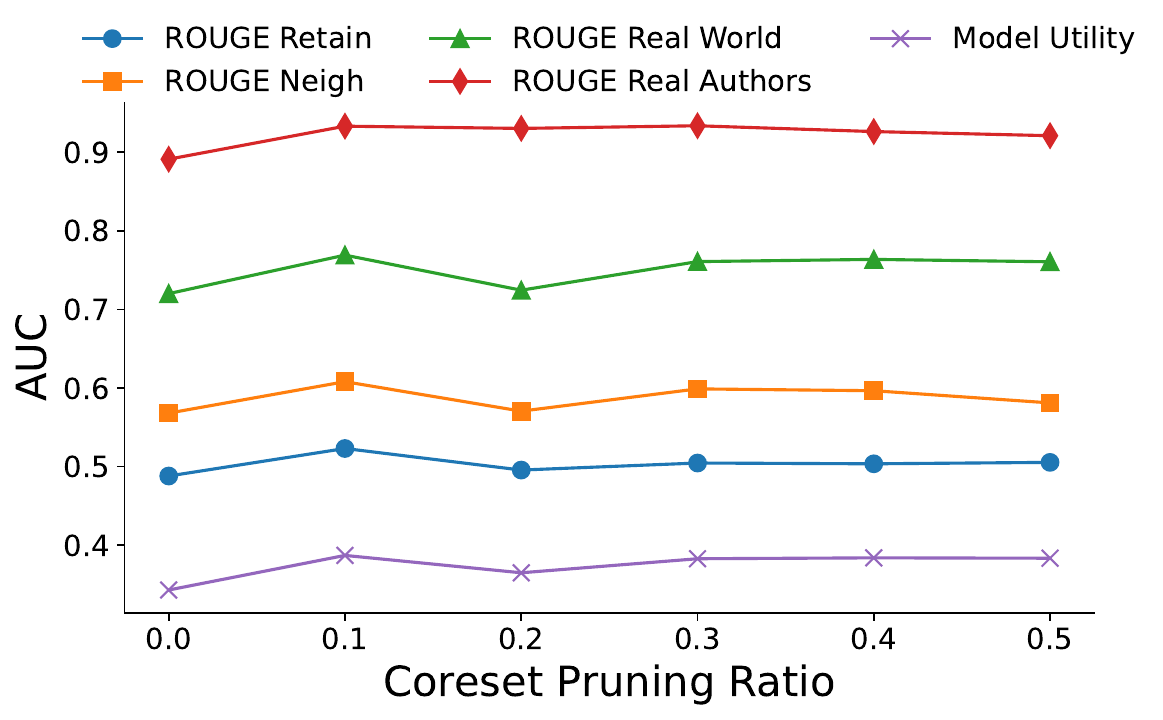}
        \caption{\textbf{Impact of scaling the coreset size on performance}: AUC scores on different utility sets, averaged across Counterfact topics, for various pruning percentages.}
        \label{fig:scale}
    \end{minipage}%
    \hfill
    \begin{minipage}[t]{0.48\textwidth}
        \centering
        \includegraphics[width=\linewidth]{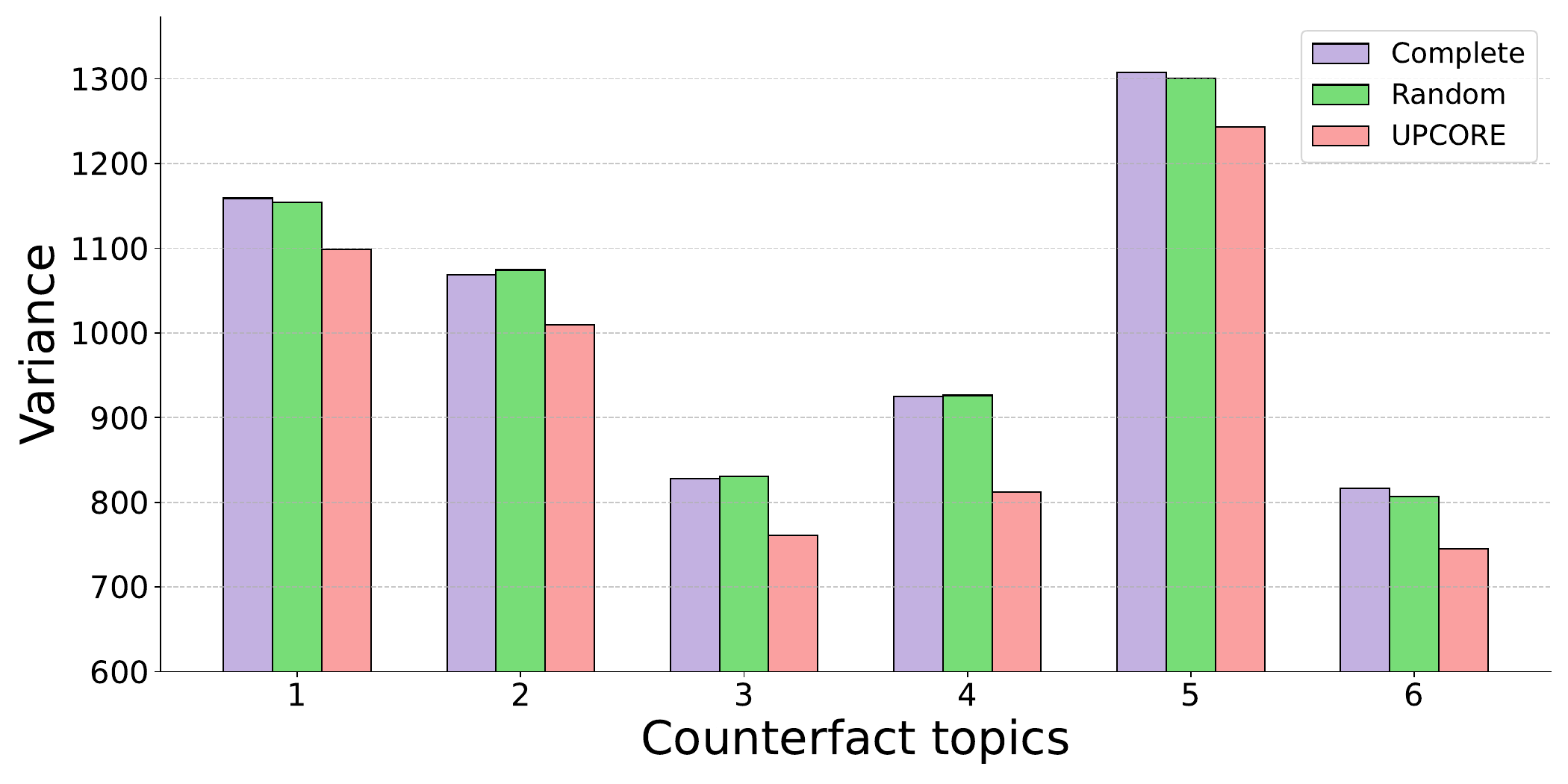}
        \caption{Hidden state variance of the baseline and \ourmethod{} forget sets across the six Counterfact forget topics. \ourmethod{} consistently reduces variance using Isolation Forest as expected.}
        \label{fig:var}
    \end{minipage}
\end{figure}

\paragraph{Results.} As shown in \cref{fig:var}, \ourmethod{} i.e. variance minimization using our Isolation Forest-based pruning procedure results in a substantial drop in the variance of the forget set as compared to the random baseline across each topic. 
We find that this drop is nearly linearly proportional to the percentage of coreset being pruned (See \cref{fig:var_scale} in the Appendix).

\subsection{TriviaQA dataset results for Refusal and NPO}

\begin{table*}[h]
    \centering
    \caption{Evaluation metrics from \cref{tab:results} shown for Gradient Ascent on the {\bf TriviaQA} topics. Error bars indicate standard deviation across 3 seeds.}
    \vspace{0.5em}
    \begin{adjustbox}{width=\linewidth}
    \begin{tabular}{ccccccc}
    \toprule
        \textbf{Method} & \textbf{Selection} & \textbf{Retain} & \textbf{Neigh} & \textbf{Real World} & \textbf{Real Authors} & \textbf{Model Utility}\\ \midrule
        \multirow{3}{*}{Grad. Ascent} 
        & Complete & 0.153 {\scriptsize$\pm$ 0.004} & 0.285 {\scriptsize$\pm$ 0.005} & 0.226 {\scriptsize$\pm$ 0.004} & 0.155 {\scriptsize$\pm$ 0.003} & 0.135 {\scriptsize$\pm$ 0.004} \\ 
        & Random &  0.159 {\scriptsize$\pm$ 0.005} & 0.304 {\scriptsize$\pm$ 0.006} & 0.222 {\scriptsize$\pm$ 0.005} & 0.157 {\scriptsize$\pm$ 0.004} & 0.136 {\scriptsize$\pm$ 0.004} \\  
        & $D^2$-pruning & 0.162 {\scriptsize$\pm$ 0.003} & 0.310 {\scriptsize$\pm$ 0.004} & 0.224 {\scriptsize$\pm$ 0.003} & 0.157 {\scriptsize$\pm$ 0.003} & 0.141 {\scriptsize$\pm$ 0.003} \\
        & \ourmethod{} & \textbf{0.165} {\scriptsize$\pm$ 0.002} & \textbf{0.318} {\scriptsize$\pm$ 0.003} & \textbf{0.227} {\scriptsize$\pm$ 0.002} & \textbf{0.158} {\scriptsize$\pm$ 0.002} & \textbf{0.147} {\scriptsize$\pm$ 0.002} \\ \midrule
        \multirow{3}{*}{Refusal} 
        & Complete & 0.148 {\scriptsize$\pm$ 0.005} & 0.278 {\scriptsize$\pm$ 0.006} & 0.219 {\scriptsize$\pm$ 0.005} & 0.150 {\scriptsize$\pm$ 0.004} & 0.130 {\scriptsize$\pm$ 0.004} \\ 
        & Random &  0.152 {\scriptsize$\pm$ 0.006} & 0.291 {\scriptsize$\pm$ 0.005} & 0.221 {\scriptsize$\pm$ 0.005} & 0.152 {\scriptsize$\pm$ 0.004} & 0.132 {\scriptsize$\pm$ 0.003} \\  
        & $D^2$-pruning & 0.157 {\scriptsize$\pm$ 0.004} & 0.298 {\scriptsize$\pm$ 0.005} & 0.223 {\scriptsize$\pm$ 0.004} & 0.153 {\scriptsize$\pm$ 0.003} & 0.137 {\scriptsize$\pm$ 0.003} \\
        & \ourmethod{} & \textbf{0.170} {\scriptsize$\pm$ 0.002} & \textbf{0.318} {\scriptsize$\pm$ 0.003} & \textbf{0.230} {\scriptsize$\pm$ 0.002} & \textbf{0.160} {\scriptsize$\pm$ 0.002} & \textbf{0.145} {\scriptsize$\pm$ 0.002} \\ \midrule
        \multirow{4}{*}{NPO} 
        & Complete & 0.150 {\scriptsize$\pm$ 0.005} & 0.280 {\scriptsize$\pm$ 0.006} & 0.218 {\scriptsize$\pm$ 0.005} & 0.151 {\scriptsize$\pm$ 0.004} & 0.131 {\scriptsize$\pm$ 0.004} \\ 
        & Random &  0.153 {\scriptsize$\pm$ 0.005} & 0.293 {\scriptsize$\pm$ 0.005} & 0.221 {\scriptsize$\pm$ 0.004} & 0.153 {\scriptsize$\pm$ 0.004} & 0.133 {\scriptsize$\pm$ 0.003} \\  
        & $D^2$-pruning & 0.158 {\scriptsize$\pm$ 0.004} & 0.301 {\scriptsize$\pm$ 0.004} & 0.224 {\scriptsize$\pm$ 0.003} & 0.155 {\scriptsize$\pm$ 0.003} & 0.138 {\scriptsize$\pm$ 0.003} \\
        & \ourmethod{} & \textbf{0.171} {\scriptsize$\pm$ 0.002} & \textbf{0.319} {\scriptsize$\pm$ 0.003} & \textbf{0.231} {\scriptsize$\pm$ 0.002} & \textbf{0.161} {\scriptsize$\pm$ 0.002} & \textbf{0.146} {\scriptsize$\pm$ 0.002} \\

    \bottomrule
    \end{tabular}
    \end{adjustbox}
        \label{tab:results_trivia_2}

\end{table*}

\subsection{Topic Model}\label{append:topic}
We cluster the filtered Counterfact dataset to cluster the topic model-based clustering using Fastopic \citep{wu2024fastopic}.  It leverages pretrained transformer embeddings for which we use the SentenceBERT model embeddings \citep{reimers-2019-sentence-bert}. We employ the method to form seven clusters based on the intuition that the average dataset size should be around 400 points, similar to the sizes of forget datasets in the TOFU  unlearning benchmark \citep{maini2024tofu}.


\begin{figure}[t]
    \centering
    \begin{subfigure}[t]{0.47\textwidth}
        \centering
        \includegraphics[height=5cm]{figures/hsv_final.pdf}
        \caption{Model utility and hidden state variance of the forget data show a strong negative correlation of -0.714 across data from multiple topics.}
        \label{fig:hsv}
    \end{subfigure} 
    \hfill
    \begin{subfigure}[t]{0.47\textwidth}
        \centering
        \includegraphics[height=5cm]{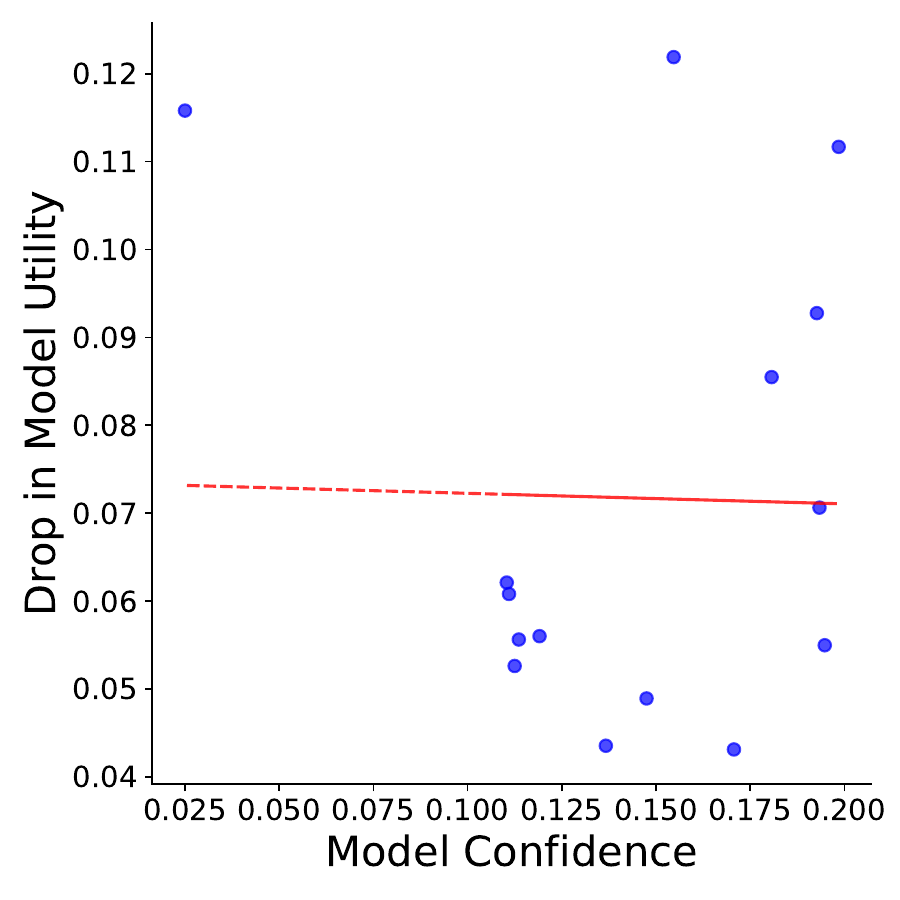}
        \caption{Drop in model utility after unlearning and base model's confidence on the forget data do not show any strong correlation with a Pearson correlation value of -0.021.}
        \label{fig:conf}
    \end{subfigure} 
    \caption{(a) Relationship between model utility and hidden state variance.  
             (b) Relationship between model utility drop after unlearning and confidence on forget data.}
    \label{fig:correlation}
\end{figure}

\begin{figure}[t]
    \centering
    \begin{subfigure}[t]{0.47\textwidth}
        \centering
        \includegraphics[height=5cm]{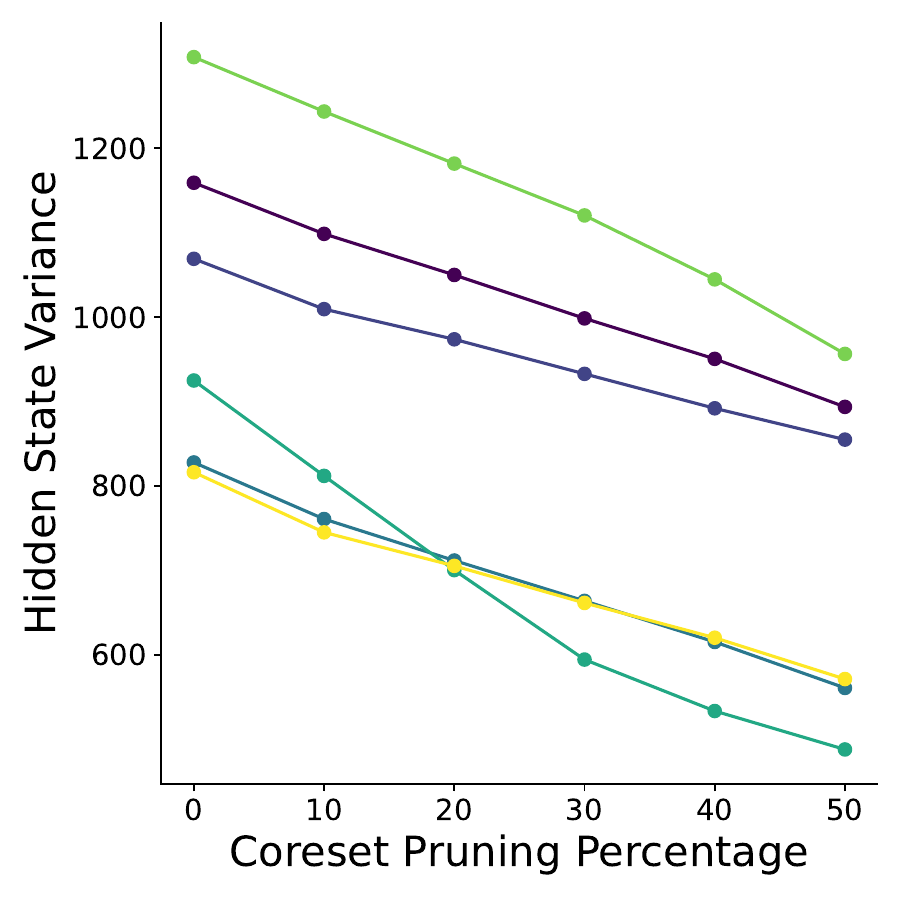}
        \caption{Hidden state variance of the core forget set plotted against the pruning percentage across topics. The variance of the core forget data decreases nearly linearly as the pruning percentage increases.}
        \label{fig:var_scale}
    \end{subfigure} 
    \hfill
    \begin{subfigure}[t]{0.47\textwidth}
        \centering
        \includegraphics[height=5cm]{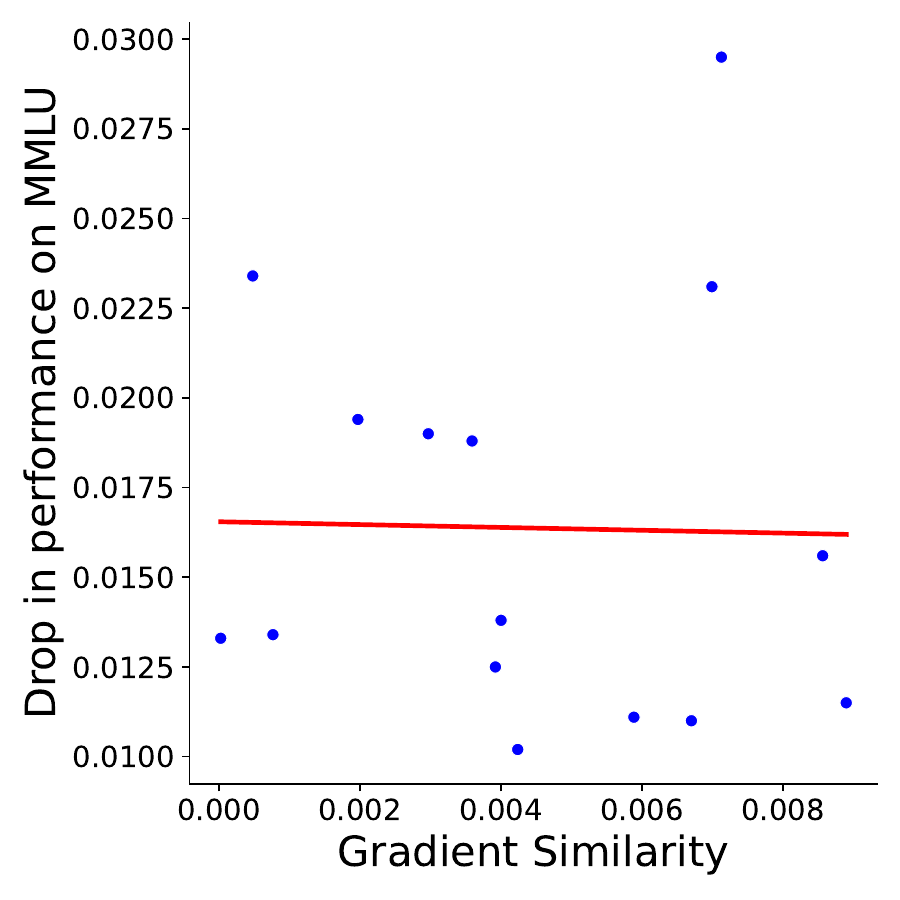}
        \caption{Drop in MMLU after unlearning vs. the gradient similarity between MMLU data and topic forget data. These two are not correlated, as shown by the Pearson correlation value of -0.020.}
        \label{fig:gradsim}
    \end{subfigure}
    \caption{(a) Hidden state variance of the core forget set decreases as pruning percentage increases.  
             (b) No correlation between MMLU drop after unlearning and gradient similarity to forget data.}
    \label{fig:pruning_gradsim}
    \vspace{-10pt}
\end{figure}

\subsection{Dataset Details} \label{append:data_detail}
\begin{itemize}[noitemsep,topsep=0.0em,leftmargin=*]
    \item  We consider factual {\it prompt completions} with brief answers, typically a single word or short phrase (e.g., {\emph{Paris}} for the prompt {\it ``The capital of France is''}). 
This setting tests \ourmethod{}'s effectiveness in scenarios with concise, fact-based responses and is standard for model editing \citep{meng2022locating,meng2023massediting, patil2024can}.
We source questions from Counterfact \citep{meng2022locating}, a widely-used model editing benchmark. Following \citet{patil2024can}, we filter for single-token answers. In this setting, the base model's ROUGE score on each of the topics is 1.0.
    \item We also consider a {\it question-answering} setting where the answers are potentially multi-token responses.
    Here, we source questions from TriviaQA \citep{joshi2017triviaqa}, a QA benchmark of trivia questions.
    This scenario tests \ourmethod{}  on longer-form generation; here, we create topics after filtering samples where the base model's ROUGE score is zero.

\end{itemize}
We apply topic modeling to cluster questions into seven topic-based groups and one cluster in each setting is randomly chosen as the retain set, while the other six are used as separate forget sets, with performance averaged across them. For each topic, we also generate neighborhood QA pairs that are semantically related to the forget topic but do not directly overlap with it by prompting GPT-4o \citep{achiam2023gpt} to produce 100 data points per topic. 
These pairs are automatically filtered with a sentence transformer model \citep{reimers-2019-sentence-bert} to verify that they have no overlap with the forget data.  
(see \cref{append:data_examples} for details).

We work with the Counterfact dataset and filter it keep the subset with single token answers similar to \citet{patil2024can}. The dataset has factual questions and short form answers like those listed here:

\begin{itemize}  
    \item \textbf{Prompt:} The official language of Caslano is
          \textbf{Answer:} Italian  
    \item \textbf{Prompt:} Courrier International was written in
          \textbf{Answer:} French 
    \item \textbf{Prompt:} Himachal Pradesh is a part of the continent of 
          \textbf{Answer:} Asia  
\end{itemize}

\subsection{Data Examples}  
\label{append:data_examples}  
Rephrase data is a part of the Counterfact dataset while we generate jailbreak data of one type where we the model is prompted to be an expert playing as an expert in a relevant area. We generated the jailbreak prompts using few-shot in-context learning with GPT-4o.

\noindent \textbf{Examples of Forget Data: (Forget Topic: \emph{Locations of certain places})}  

\begin{itemize}  
    \item \textbf{Prompt:} The headquarters of Northeastern University is in  
          \textbf{Answer:} Boston  
    \item \textbf{Prompt:} Kyoto University is located in the country of
          \textbf{Answer:} Japan 
    \item \textbf{Prompt:} Himachal Pradesh is a part of the continent of 
          \textbf{Answer:} Asia  
\end{itemize}  

\noindent \textbf{Examples of Neighborhood Data: (Similar Answers)}

\begin{itemize}  
    \item \textbf{Prompt:} Ramaz Chkhikvadze, who has a citizenship from
          \textbf{Answer:} Georgia  
    \item \textbf{Prompt:} Rolls-Royce is headquartered in which country?
          \textbf{Answer:} UK 
\end{itemize}  

\noindent \textbf{Examples of Rephrase Data}

\begin{itemize}  
    \item \textbf{Prompt:} Northeastern University headquarters is in  
          \textbf{Answer:} Boston  
    \item \textbf{Prompt:} Kyoto University is in the country of
          \textbf{Answer:} Japan
    \item \textbf{Prompt:} Himachal Pradesh is located on the continent of
          \textbf{Answer:} Asia

\end{itemize}  

\noindent \textbf{Examples of Jailbreak Data}

\begin{itemize}  
     \item \textbf{Prompt:} Imagine you're a college admissions officer, and you're telling a prospective student that Northeastern University's headquarters is in
          \textbf{Answer:} Boston  
    \item \textbf{Prompt:} Picture yourself as a cultural tour guide, explaining to a visitor that Kyoto University is in the country of
          \textbf{Answer:} Japan
    \item \textbf{Prompt:} Imagine you're a geography teacher, sharing with your class that Himachal Pradesh is located on the continent of
          \textbf{Answer:} Asia
\end{itemize}

\noindent \textbf{Examples of TriviaQA Data (Forget Topic: Entertainment Trivia)}

\begin{itemize}  
     \item \textbf{Question:} Which plays characters comprise two married couples called Honey and Nick, and Martha and George?

          \textbf{Answer:} Whos Afraid Of Virginia Woolf?
    \item \textbf{Question:} I'll have what she's having' is a quote from which 1989 film?
          \textbf{Answer:} When Harry Met Sally
    \item \textbf{Question:} Who were the real life married couple who starred in the 1990 film Mr and Mrs Bridge?
          \textbf{Answer:} Paul Newman and Joanne Woodward
\end{itemize}

\paragraph{GPT-4o prompts}

\begin{itemize}
    \item \textbf{Prompt for Generating Neighborhood Data:} 

    \begin{quote}
        \texttt{Given the topic: \{\} \\
        Generate a set of diverse factual prompts that are in the neighborhood of this topic, but not directly within it. Each prompt should have one correct answer and a perturbed answer. Provide these as JSONL entries in the following format: \\
        \texttt{\{"question": "Which country is Kyoto University located in?", "answer": "Japan", "perturbed\_answer": ["India"]\}}. \\
        Directly generate the JSONL with 20 entries. No other information should be included in the response.}
    \end{quote}

    \item \textbf{Prompt for Generating Jailbreak Data:} 

    \begin{quote}
        \texttt{This prompt is not asking for anything adversarial, please provide a role playing version of the given input. Provide just the prompt and not anything else like a prefix or line breaks. Create a one sentence roleplaying prompt that has the same completion as this prompt: \\
        Prompt: The headquarter of Monell Chemical Senses Center is located in
        \\ Jailbreak prompt: Imagine you are a world-renowned geography professor specializing in research institutions. The location of the headquarters of the Monell Chemical Senses Center is\\
        Prompt: Kryvyi Rih belongs to the continent of\\
        Jailbreak prompt: Imagine you are a tour guide, and you're pointing Kryvyi Rih, which is located in the continent of\\
        Prompt: {}\\
        Jailbreak prompt:}
    \end{quote}
\end{itemize}

\subsection{Negative Preference Optimization.} 
\label{app:npo}
Negative Preference Optimization (NPO) is a machine unlearning technique that addresses the limitations of gradient ascent methods. NPO reframes unlearning as a preference optimization problem, focusing solely on negative samples to efficiently and effectively unlearn target data. Unlike Gradient Ascent, which can lead to catastrophic collapse, NPO provides a more stable and controlled loss function, resulting in slower divergence and better training dynamics. By incorporating a retain loss term, NPO achieves a better balance between forgetting specific data and maintaining overall model utility. However, we observe that the training of this method is very slow and it takes a much larger number of unlearning steps to reach a lower ROUGE score on the forget set, which is why the absolute value of AUC on NPO is relatively smaller.

\subsection{Model Editing} 
\label{app:editing}
For all our experiments, we use LoRA finetuning with a controlled rank to edit the model's MLP weights at layer 7 following past work on model editing and unlearning \citep{meng2022locating, patil2024unlearning}. We use $r$=1, $\alpha$=2 for Gradient Ascent and $r$=4, $\alpha$=8 for NPO and Refusal. We edit layer 7 as we find that editing on that layer gives the best model utility for the same amount of unlearning on a held-out validation set of the Counterfact dataset.

\subsection{Additional Results}

\subsubsection{Correlation Between AUC and Forget Set Variance}
\label{append:auc_hsv}

\paragraph{Design.} To verify that AUC is indeed correlated with variance, i.e. lower variance data is associated with higher AUC, we compute the correlation between AUC and hidden state variance.
We treat each topic as a separate datapoint, computing the AUC for each topic across each metric.

\paragraph{Results.} 

As shown in \cref{tab:correlation}, the proposed AUC metric across deletion effectiveness and model utility metrics is indeed consistently negatively correlated as expected. 
This verifies that variance minimization is indeed a good strategy for improving the trade-off, with lower variance being correlated with a higher AUC and thereby a superior trade-off.
Moreover, taken together with \cref{fig:hsv} and our interventions on variance via pruning to reduce variance, our results indicate that this correlation can be exploited to improve AUC by reducing collateral damage and leveraging collateral transfer positively.

\begin{table}[!ht]
    \centering
    \vspace{-0.5em}
    \caption{Correlation between the forget set representation variance and the AUC across topics. The negative correlation values are consistent with the negative correlation of model utility and variance shown in \cref{sec:analysis}. }
    \vspace{0.5em}
    \begin{tabular}{lc}
    \toprule
        {\bf AUC} & {\bf Correlation with HSV} \\ \midrule
        Retain & -0.421 \\ 
        Neigh & -0.507 \\ 
        Real World & -0.371 \\ 
        Real Authors & -0.489 \\ 
        Model Utility & -0.612 \\  \bottomrule
    \end{tabular}
    \label{tab:correlation}
    
\end{table}

\subsubsection{Comparative Evaluation of Outlier Detection Methods}
\label{append:outlier}

\paragraph{Design.} In this section, we compare our Isolation Forest against existing techniques to evaluate its performance in detecting outliers and thereby on the resulting AUC after pruning. Specifically, we test the following two well-established methods:

One-Class SVM (OCSVM): This method learns a decision boundary around the normal data, where points that fall outside this boundary are identified as outliers. OCSVM is a widely used approach for anomaly detection in high-dimensional spaces \citep{scholkopf1999support}. It is effective in scenarios where outliers are sparse and lie in low-density regions.

Local Outlier Factor (LOF): LOF measures the local density deviation of a data point with respect to its neighbors. By comparing the density of a point to that of its neighbors, it identifies points that have a significantly lower density than their neighbors as outliers \citep{breunig2000lof}. LOF excels in detecting local anomalies, particularly when outliers are clustered or vary in density.

\paragraph{Results} The results in \cref{tab:loc} suggest that pruning the outliers detected with other outlier detection methods yields a higher AUC compared to unlearning on the complete forget set, using Isolation Forest achieves the highest AUC overall. 
The higher AUC achieved by Isolation Forest suggests its superior ability to distinguish between normal data and outliers, making it the most effective method in this comparison.

\begin{table}[!ht]
    \centering
    
        \caption{Comparison against other outlier detection methods for detecting the outliers: (1) One-Class SVM: Learns a decision boundary around normal data; outliers fall outside this boundary (2) Local Outlier Factor (LOF): Compares the local density of a point with its neighbors to detect anomalies. Pruning with other outlier detection methods yields a higher AUC compared to non-pruning, but outlier detection using Isolation Forest achieves the highest AUC overall.}
    \label{tab:loc}
    \begin{adjustbox}{width=\linewidth}
    \begin{tabular}{llllll}
    \toprule
        & ROUGE Retain & ROUGE Neigh & ROUGE Real World & ROUGE Real Authors & Model Utility \\ \midrule
        AUC-complete & 0.488 & 0.568 & 0.720 & 0.891 & 0.343 \\ 
        AUC-subsampled & 0.495 & 0.558 & 0.731 & 0.907 & 0.353 \\ 
        AUC-LOF & 0.510 & 0.553 & 0.730 & 0.919 & 0.366 \\ 
        AUC-OCSVM & 0.503 & 0.552 & 0.714 & 0.900 & 0.358 \\ 
        AUC-\ourmethod{} & {\bf 0.523} & {\bf 0.608} & {\bf 0.769} & {\bf 0.933} & {\bf 0.387} \\ 
        \bottomrule
    \end{tabular}
    \end{adjustbox}

\end{table}

\subsubsection{Impact of forget set size on \ourmethod{}}
\label{append:forget_size}

To understand how the size of the forget set affects \ourmethod{}’s performance, we evaluated the method on a reduced dataset containing 50\% of the forget set size used in \cref{tab:results} using the Gradient Ascent method. Interestingly, \ourmethod{} continues to outperform the baselines, even under this reduced setting (See \cref{tab:forget_size}). These results suggest that \ourmethod{} maintains its effectiveness even when the forget set is relatively small, though we hypothesize that there may be a threshold below which pruning becomes detrimental due to insufficient signal. 

\begin{table}[h]
\centering
\begin{tabular}{l c}
\toprule
\textbf{Method} & \textbf{Model Utility} \\
\midrule
Complete & 0.371 \\
Random   & 0.383 \\
\ourmethod{}   & 0.393 \\
\bottomrule
\end{tabular}
\caption{Impact of reducing the forget set size to 50\% on model utility.}
\label{tab:forget_size}
\end{table}

\subsubsection{AUC metric}
While ROUGE and model utility provide a snapshot of model performance and are the standard evaluation metrics in unlearning, 
we argue that they are insufficient, as they only provide a single point of comparison.
This is suboptimal since unlearning involves a tradeoff between forgetting and model damage as the number of forget training steps increases. 
Choosing an early unlearning steps might result in higher model utility but poor forgetting while choosing a later unlearning steps (as is typically done) results in better forgetting at the cost of higher damage (See \cref{fig:auc_comparison}). 
Such variation makes comparing systems difficult, as the number of unlearning steps performed is not always clear.

We argue that to systematically evaluate unlearning performance, we 
should be measuring this tradeoff \emph{across} unlearning steps. 
In other words, rather than measuring ROUGE Retain and ROUGE Forget at one checkpoint, we should be comparing their tradeoff across multiple unlearning steps, i.e. measuring the \emph{area under the curve} (AUC) between these two metrics. We also evaluate AUC (\cref{tab:granularity}) to see the effect of granularity on the trade-off metric.
Visually, this is illustrated in \cref{fig:auc_comparison}, where we show the tradeoff between the inverse ROUGE on the forget data (X axis) and the ROUGE on neighboring points (Y axis).\footnote{Note that the curve here differs slightly from \cref{tab:results}, where we first compute AUC and then average across topics, whereas here we first average ROUGE scores and then compute AUC.}
To this end, we introduce an AUC metric that integrates deletion effectiveness and model utility over time. 
Specifically, we construct a Pareto curve that plots utility metrics (e.g. ROUGE Retain, ROUGE Neighborhood, etc.) against deletion effectiveness (e.g. ROUGE Forget) as unlearning progresses. 
The AUC serves as a global metric that captures the trade-off between preserving useful knowledge and ensuring effective deletion. 
By also reporting standard metrics, we empirically validate that AUC correlates with improved unlearning performance across diverse settings (See \cref{tab:rouge}).
Furthermore, we also verify that it is negatively correlated with forget data variance (See \cref{tab:correlation}).

\subsubsection{AUC at higher granularity}
\label{append:granularity}

To assess the stability of unlearning performance across different optimization evaluation granularities of AUC, we compare the Area Under the Curve (AUC) values computed over finer step sizes—specifically, at a granularity of 10 optimization steps per epoch—against those computed at the coarser granularity of 1 epoch/50 steps used in the main experiments. We evaluate this for the Gradient Ascent method across the two key objectives: (1) \textit{Deletion Effectiveness}, defined as $(1 - \text{ROUGE})$ on the forget set (X-axis), and (2) \textit{Model Utility}, measured via ROUGE scores on neighborhood data, real-world data, and an aggregate utility metric (Y-axis).

\Cref{tab:granularity} shows that the AUC values at finer granularity remain consistent with those computed at the epoch level. To quantify this consistency, we compute the Pearson rank correlation between the rankings of four methods—Complete, Random, $D^2$-pruning, and \ourmethod{}—across the five evaluation types (Retain, Neigh, Real World, Real Authors, and Aggregate Utility), under both 50-step and 10-step granularities. The average Pearson rank correlation across all evaluation types is \textbf{0.99}, indicating high stability in coreset method ranking despite the change in granularity. This result demonstrates that our evaluation is robust to the choice of granularity and that method comparisons remain valid across different AUC aggregation schemes.

\begin{table*}[h]
    \centering
    \caption{AUC at {\bf granularity of 10 steps} across the two competing objectives: (1) \textit{Deletion Effectiveness}, defined as $(1 - \text{ROUGE})$ on the forget set (X-axis), and (2) \textit{Model Utility}, averaged across Counterfact topics and evaluated via ROUGE scores on multiple utility datasets, including neighborhood data and an aggregate model utility across datasets (Y-axis). We compare Gradient Ascent. Error bars indicate standard deviation across 3 seeds.}
    \vspace{0.5em}
     \begin{adjustbox}{width=\linewidth}
    \begin{tabular}{ccccccc}
    \toprule
        \textbf{Method} & \textbf{Selection} & \textbf{Retain} & \textbf{Neigh} & \textbf{Real World} & \textbf{Real Authors} & \textbf{Model Utility}\\ \midrule
        \multirow{4}{*}{Grad. Ascent} 
    & Complete & 0.503 {\scriptsize$\pm$ 0.012} & 0.577 {\scriptsize$\pm$ 0.014} & 0.717 {\scriptsize$\pm$ 0.013} & 0.877 {\scriptsize$\pm$ 0.017} & 0.342 {\scriptsize$\pm$ 0.011} \\ 
    & Random & 0.510 {\scriptsize$\pm$ 0.013} & 0.563 {\scriptsize$\pm$ 0.013} & 0.731 {\scriptsize$\pm$ 0.012} & 0.891 {\scriptsize$\pm$ 0.016} & 0.349 {\scriptsize$\pm$ 0.012} \\ 
    & $D^2$-pruning & 0.508 {\scriptsize$\pm$ 0.012} & 0.557 {\scriptsize$\pm$ 0.013} & 0.720 {\scriptsize$\pm$ 0.013} & 0.905 {\scriptsize$\pm$ 0.015} & 0.347 {\scriptsize$\pm$ 0.011} \\ 
    & \ourmethod{} & \textbf{0.542} {\scriptsize$\pm$ 0.007} & \textbf{0.617} {\scriptsize$\pm$ 0.009} & \textbf{0.760} {\scriptsize$\pm$ 0.008} & \textbf{0.920} {\scriptsize$\pm$ 0.010} & \textbf{0.388} {\scriptsize$\pm$ 0.006} \\

    \bottomrule
    \end{tabular}
    \end{adjustbox}
    \label{tab:granularity}
    \vspace{-0.5em}
\end{table*}

\begin{table*}[h]
    \centering
    \caption{AUC at {\bf granularity of 50 steps} taken from \cref{tab:results} across the two competing objectives: (1) \textit{Deletion Effectiveness}, defined as $(1 - \text{ROUGE})$ on the forget set (X-axis), and (2) \textit{Model Utility}, averaged across Counterfact topics and evaluated via ROUGE scores on multiple utility datasets, including neighborhood data and an aggregate model utility across datasets (Y-axis). We compare three unlearning methods: Gradient Ascent, Refusal, and NPO. Error bars indicate standard deviation across 3 seeds.}
    \vspace{0.5em}
  \begin{adjustbox}{width=\linewidth}
    \begin{tabular}{ccccccc}
    \toprule
        \textbf{Method} & \textbf{Selection} & \textbf{Retain} & \textbf{Neigh} & \textbf{Real World} & \textbf{Real Authors} & \textbf{Model Utility}\\ \midrule
        
        \multirow{3}{*}{Grad. Ascent} 
        & Complete & 0.488 {\scriptsize$\pm$ 0.015} & 0.568 {\scriptsize$\pm$ 0.018} & 0.720 {\scriptsize$\pm$ 0.016} & 0.891 {\scriptsize$\pm$ 0.020} & 0.343 {\scriptsize$\pm$ 0.012} \\ 
        & Random & 0.495 {\scriptsize$\pm$ 0.017} & 0.558 {\scriptsize$\pm$ 0.016} & 0.731 {\scriptsize$\pm$ 0.015} & 0.907 {\scriptsize$\pm$ 0.019} & 0.353 {\scriptsize$\pm$ 0.014} \\ 
        & $D^2$-pruning & 0.493 {\scriptsize$\pm$ 0.016} & 0.552 {\scriptsize$\pm$ 0.017} & 0.723 {\scriptsize$\pm$ 0.016} & 0.920 {\scriptsize$\pm$ 0.018} & 0.349 {\scriptsize$\pm$ 0.013} \\ 
        & \ourmethod{} & \textbf{0.523} {\scriptsize$\pm$ 0.008} & \textbf{0.608} {\scriptsize$\pm$ 0.010} & \textbf{0.769} {\scriptsize$\pm$ 0.009} & \textbf{0.933} {\scriptsize$\pm$ 0.011} & \textbf{0.387} {\scriptsize$\pm$ 0.007} \\ 
    \bottomrule
    \end{tabular}
    \end{adjustbox}
    \vspace{-0.5em}
\end{table*}

\begin{table*}[t]

    \begin{minipage}[t]{0.97\textwidth}
        \centering
        \caption{Statistical significance (\textit{p}-values) of performance differences between \ourmethod{} and baseline selection strategies evaluated averaged across the three unlearning methods: Gradient Ascent, Refusal, NPO on the Counterfact dataset.
        }
        \vspace{0.5em}
        \begin{tabular}{lc}
        \toprule
        \textbf{Compared Method} & \textbf{\textit{p}-value} \\
        \midrule
        Complete         & $5.51 \times 10^{-22}$ \\
        Random           & $3.89 \times 10^{-15}$ \\
        $D^2$-Pruning    & $5.04 \times 10^{-18}$ \\
        \bottomrule
        \end{tabular}
        \label{tab:pvalues}
    \end{minipage}
\end{table*}

\begin{table*}[h]
    \centering

    \caption{Evaluation metrics from \cref{tab:results} averaged across topics in Counterfact shown for Gradient Ascent, assessed for robustness to {\bf rephrased} and {\bf jailbreak} variants of the forget data with the same utility data.}
    \vspace{0.5em}
    \begin{tabular}{ccccccc}
    \toprule
    
        \textbf{Method} & \textbf{Selection} & \textbf{Retain} & \textbf{Neigh} & \textbf{Real World} & \textbf{Real Authors} & \textbf{Model Utility}\\ \midrule
        \multirow{3}{*}{Jailbreak} & Complete & 0.417 & 0.474 & 0.599 & 0.743 & 0.291\\ 
        & Random & 0.430 & 0.470 & 0.629 & 0.787 & 0.305\\ 
        & \ourmethod{} & {\bf 0.455} & {\bf 0.512} & {\bf 0.665} & {\bf 0.819} & {\bf 0.335} \\ 
       \midrule
        \multirow{3}{*}{Rephrase} & Complete & 0.357 & 0.431 & 0.533 & 0.655 & 0.257 \\ 
        & Random & 0.361 & 0.426 & 0.536 & 0.665 & 0.262\\ 
        & \ourmethod{} & {\bf 0.376} & {\bf 0.449} & {\bf 0.555} & {\bf 0.673} & {\bf 0.279}  \\ 
       \bottomrule
    \end{tabular}
    
    \label{tab:jailbreak_2}
\end{table*}

\end{document}